\documentclass{ws-jmrr}
\usepackage[sort, compress, super]{cite}
\usepackage{etoolbox} 
\usepackage{algorithm}
\usepackage[noend]{algpseudocode}
\usepackage{xspace}    
\usepackage{upgreek}   
\usepackage{booktabs}  
\usepackage{tikz}
\usepackage{paralist}
\usepackage{graphicx}
\usepackage[hyphens,spaces,obeyspaces]{url} 
\usepackage{amsmath}
\usepackage{dsfont}
\usepackage{import}    
\usepackage{markdown}
\usepackage{xfrac}     
\usepackage{multirow}
\usepackage{afterpage}
\usepackage{pbalance}  
\usepackage[
  colorlinks,
  urlcolor=black,
  linkcolor=black,
  citecolor=black,
  ]{hyperref}
\usepackage{bm} 
\usepackage{suffix}     
\usepackage[binary-units=true]{siunitx}[=v2]
\usepackage[normalem]{ulem} 

\makeatletter
\def\catchline#1#2#3#4#5{
    \clcount=#3
\expandafter\def\expandafter\@clinebuf\expandafter
  {\@clinebuf\catchlinefont
    \noindent Preprint of an article submitted for consideration in Journal of Medical Robotics Research\par
    \noindent \copyright\ #3 World Scientific Publishing Company \url{https://www.worldscientific.com/worldscinet/jmrr}.\par
    \noindent\vskip-\baselineskip \hphantom{#4 \hskip2em #5}
  }\relax\par
}%
\makeatother

\catchline{0}{0}{2022}{}{}

\makeatletter
\appto{\input@path}{{inputs/}}
\makeatother

\newbool{enablecomments}
\setbool{enablecomments}{true}

\DeclareSIUnit[number-unit-product = ]\percent{\%} 
\DeclareSIUnit\rev{rev} 
\newcommand\numquart[3]{\num{#1} (Q: \numrange{#2}{#3})}

\newcommand{\narrowdoublestack}[2]{\shortstack{#1\\#2}}

\renewcommand{\toprule}{\Hline\\[-6.5pt]}
\renewcommand{\bottomrule}{\botrule}
\renewcommand{\midrule}{\colrule}

\makeatletter
\def\MyTabularWrap{} 
\def\MyAtTableBegin{
  \centering\parindent\z@\ignorespaces\noindent
  \NINE
  \def\MyTabularWrap{\par\setbox\tempbox\hbox}
}
\def\MyAtTableEnd{
  \tablewd\hsize\advance\tablewd-\wd\tempbox\global\divide\tablewd\tw@
  \ifdim\wd\captionbox<\wd\tempbox\centerline{\unhbox\captionbox}
  \else\leftskip\tablewd\rightskip\leftskip{\unhbox\captionbox}\par
  \fi\vskip5pt\centerline{\box\tempbox}%
  \def\MyAtTabularEnd{}
}
\makeatother
\AtBeginEnvironment{table}{\MyAtTableBegin}
\AtEndEnvironment{table}{\MyAtTableEnd}
\AtBeginEnvironment{table*}{\MyAtTableBegin}
\AtEndEnvironment{table*}{\MyAtTableEnd}
\BeforeBeginEnvironment{tabular}{\MyTabularWrap\bgroup}
\AfterEndEnvironment{tabular}{\egroup}

\renewcommand{\vec}[1]{\bm{#1}}

\newcommand{\etal}{\textit{et al.}\xspace}

\newcommand{\Model}[2]{\text{\expandafter\MakeUppercase#1-\expandafter\MakeUppercase#2}\xspace}
\newcommand{\ModelShearSingle}{\Model{shear}{single}}
\newcommand{\ModelShearDouble}{\Model{shear}{double}}
\newcommand{\ModelForceSingle}{\Model{force}{single}}
\newcommand{\ModelForceDouble}{\Model{force}{double}}

\definecolor{purple}{RGB}{210, 0, 210}
\definecolor{darkred}{RGB}{150, 0, 0}
\definecolor{greenishblue}{RGB}{37, 132, 172}
\definecolor{blueishgreen}{RGB}{0, 172, 72}
\definecolor{orange}{RGB}{255, 150, 0}
\definecolor{darkgreen}{RGB}{0, 100, 0}
\definecolor{darkblue}{RGB}{30, 30, 180}
\definecolor{darkgrey}{RGB}{130, 130, 130}

\newcommand{\missingfigure}{%
  \mylog{Missing figure}%
  \includegraphics[width=\columnwidth]{missingfig.pdf}%
}
\WithSuffix\newcommand\missingfigure*{%
  \mylog{Missing figure}%
  \includegraphics[width=\textwidth]{missingfig.pdf}%
}

\newrobustcmd{\mylog}[1]{\wlog{mylog: #1}}

\newrobustcmd{\mytodo}[3]{%
  \ifbool{enablecomments}{%
    \mylog{mytodo: TODO from #1 added}%
    {\color{#2}[#1: #3]}%
  }{
    \mylog{mytodo: TODO from #1 suppressed}%
  }%
}

\WithSuffix\newcommand\mytodo*[3]{%
  \ifbool{enablecomments}{%
    \mylog{mytodo: TODO* from #1 added}%
    {\color{#2}[#1] #3}%
  }{
    \mylog{mytodo: TODO*from #1 printed normally}%
    {#3}
  }%
}

\newcommand{\AK}[1]{\mytodo{AK}{purple}{#1}}
\newcommand{\MB}[1]{\mytodo{MB}{blueishgreen}{#1}}
\newcommand{\CR}[1]{\mytodo{CR}{orange}{#1}}
\newcommand{\OS}[1]{\mytodo{OS}{cyan!70!blue}{#1}}
\WithSuffix\newcommand\AK*[1]{\mytodo*{AK}{purple}{#1}}
\WithSuffix\newcommand\MB*[1]{\mytodo*{MB}{blueishgreen}{#1}}
\WithSuffix\newcommand\CR*[1]{\mytodo*{CR}{orange}{#1}}
\WithSuffix\newcommand\OS*[1]{\mytodo*{OS}{orange}{#1}}

\author{%
  Michael Bentley$^{a}$,
  Caleb Rucker$^{b}$,
  Chakravarthy Reddy$^{c}$,
  Oren Salzman$^{d}$,
  and Alan Kuntz$^{a}$%
}

\address{$^a$%
  Robotics Center and Kahlert School of Computing, \\
  University of Utah,
  Salt Lake City, UT 84112, USA \\
  E-mail: \href{mailto:michael.bentley@utah.edu}{michael.bentley@utah.edu}
}

\address{$^b$%
  The Department of Mechanical, Aerospace, and Biomedical Engineering, \\
  University of Tennessee,
  Knoxville, TN 37996, USA
}

\address{$^c$%
  Huntsman Cancer Institute and School of Medicine, \\
  University of Utah,
  Salt Lake City, UT 84112, USA
}

\address{$^d$%
  Department of Computer Science, \\
  Technion - Israel Institute of Technology,
  Technion City, Haifa, 3200003, Israel
}

\graphicspath{{images/}}

\title{
  Safer Motion Planning of Steerable Needles via a \\
  Shaft-to-Tissue Force Model
  }
\def\mykeywords{%
  Force modeling;
  robotic needle steering;
  bottleneck cost function;
  motion planning;
  continuum robots
}

\begin{document}

\twocolumn[
  \maketitle
  \begin{abstract}
  Steerable needles are capable of accurately targeting difficult-to-reach clinical sites in the body.
  By bending around sensitive anatomical structures, steerable needles have the potential to reduce the invasiveness of many medical procedures.
  However, inserting these needles with curved trajectories increases the risk of tissue damage due to perpendicular forces exerted on the surrounding tissue by the needle's shaft, potentially resulting in lateral shearing through tissue.
  Such forces can cause significant damage to surrounding tissue, negatively affecting patient outcomes.
  In this work, we derive a tissue and needle force model based on a Cosserat string formulation, which describes the normal forces and frictional forces along the shaft as a function of the planned needle path, friction model and parameters, and tip piercing force.
  We propose this new force model and associated cost function as a safer and more clinically relevant metric than those currently used in motion planning for steerable needles.
  We fit and validate our model through physical needle robot experiments in a gel phantom.
  We use this force model to define a bottleneck cost function for motion planning and evaluate it against the commonly used path-length cost function in hundreds of randomly generated 3-D environments.
  Plans generated with our force-based cost show a \SI{62}{\percent} reduction in the peak modeled tissue force with only a \SI{0.07}{\percent} increase in length on average compared to using the path-length cost in planning.
  Additionally, we demonstrate the ability to plan motions with our force-based cost function in a lung tumor biopsy scenario from a segmented computed tomography (CT) scan.
  By planning motions for the needle that aim to minimize the modeled needle-to-tissue force explicitly, our method plans needle paths that may reduce the risk of significant tissue damage while still reaching desired targets in the body.
\end{abstract}

  \keywords{\mykeywords}
]

  \section{Introduction}
\label{sec:intro}

%
%
%
%
%

\begin{figure}[tpb]
  \centering
  {
    \def\svgwidth{\columnwidth}
    \Large
\begingroup%
  \makeatletter%
  \providecommand\color[2][]{%
    \errmessage{(Inkscape) Color is used for the text in Inkscape, but the package 'color.sty' is not loaded}%
    \renewcommand\color[2][]{}%
  }%
  \providecommand\transparent[1]{%
    \errmessage{(Inkscape) Transparency is used (non-zero) for the text in Inkscape, but the package 'transparent.sty' is not loaded}%
    \renewcommand\transparent[1]{}%
  }%
  \providecommand\rotatebox[2]{#2}%
  \newcommand*\fsize{\dimexpr\f@size pt\relax}%
  \newcommand*\lineheight[1]{\fontsize{\fsize}{#1\fsize}\selectfont}%
  \ifx\svgwidth\undefined%
    \setlength{\unitlength}{305.74658203bp}%
    \ifx\svgscale\undefined%
      \relax%
    \else%
      \setlength{\unitlength}{\unitlength * \real{\svgscale}}%
    \fi%
  \else%
    \setlength{\unitlength}{\svgwidth}%
  \fi%
  \global\let\svgwidth\undefined%
  \global\let\svgscale\undefined%
  \makeatother%
  \begin{picture}(1,0.48085496)%
    \lineheight{1}%
    \setlength\tabcolsep{0pt}%
    \put(0,0){\includegraphics[width=\unitlength,page=1]{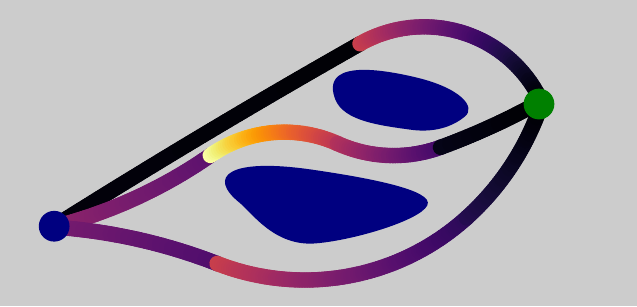}}%
    \put(0.08325537,0.04420716){\makebox(0,0)[t]{\lineheight{1.14999998}\smash{\begin{tabular}[t]{c}start\end{tabular}}}}%
    \put(0.86708021,0.3556637){\makebox(0,0)[lt]{\lineheight{1.14999998}\smash{\begin{tabular}[t]{l}goal\end{tabular}}}}%
  \end{picture}%
\endgroup%

  }
  \caption[%
    A tissue normal force profile for three example paths using our new tissue normal force model.
  ]{%
    A tissue normal force profile for three example paths using our new needle-to-tissue force and friction model.
    Darker colors indicate small values and lighter colors indicate large values.
    The upper and lower paths have similar lengths and maximum tissue forces.
    The middle path is \SI{8.5}{\percent} shorter but has \SI{79}{\percent} higher maximum normal force despite the middle path's maximum curvature matching the top path's maximum curvature.
    %
    According to our model, the middle path has a higher probability of causing significant damage to the surrounding tissue than the top or bottom paths.
    %
    }
  \label{fig:three-path-force}
\end{figure}

\begin{figure}[tpb]
  \centering
  {
    \def\svgwidth{\columnwidth}
\begingroup%
  \makeatletter%
  \providecommand\color[2][]{%
    \errmessage{(Inkscape) Color is used for the text in Inkscape, but the package 'color.sty' is not loaded}%
    \renewcommand\color[2][]{}%
  }%
  \providecommand\transparent[1]{%
    \errmessage{(Inkscape) Transparency is used (non-zero) for the text in Inkscape, but the package 'transparent.sty' is not loaded}%
    \renewcommand\transparent[1]{}%
  }%
  \providecommand\rotatebox[2]{#2}%
  \newcommand*\fsize{\dimexpr\f@size pt\relax}%
  \newcommand*\lineheight[1]{\fontsize{\fsize}{#1\fsize}\selectfont}%
  \ifx\svgwidth\undefined%
    \setlength{\unitlength}{252.00000357bp}%
    \ifx\svgscale\undefined%
      \relax%
    \else%
      \setlength{\unitlength}{\unitlength * \real{\svgscale}}%
    \fi%
  \else%
    \setlength{\unitlength}{\svgwidth}%
  \fi%
  \global\let\svgwidth\undefined%
  \global\let\svgscale\undefined%
  \makeatother%
  \begin{picture}(1,0.90199538)%
    \lineheight{1}%
    \setlength\tabcolsep{0pt}%
    \put(0,0){\includegraphics[width=\unitlength,page=1]{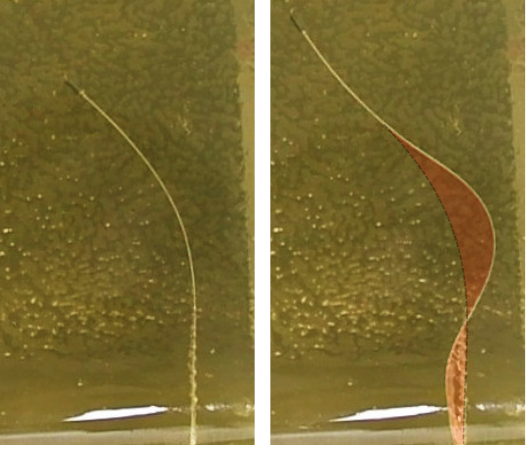}}%
    \put(0.24241394,0.01018414){\color[rgb]{0,0,0}\makebox(0,0)[t]{\lineheight{1.25}\smash{\begin{tabular}[t]{c}(a)\end{tabular}}}}%
    \put(0.75758603,0.01018414){\color[rgb]{0,0,0}\makebox(0,0)[t]{\lineheight{1.25}\smash{\begin{tabular}[t]{c}(b)\end{tabular}}}}%
  \end{picture}%
\endgroup%

  }
  \caption[%
    An example of a needle (a)~before and (b)~after shearing through a gelatin phantom.
  ]{%
    An example of a needle \textbf{(a)}~before and \textbf{(b)}~after shearing through a gelatin phantom.
    Shearing occurs when the force applied by the needle's shaft to the surrounding tissue is larger than the tissue can support without fracturing, causing the needle shaft to slice laterally through tissue.
    In (b) we show a dotted line of the shape prior to shearing, and a red region showing the sheared area.
    This figure demonstrates significant tissue damage due to shearing in a gel-based tissue phantom.
    }
  \label{fig:shearing}
\end{figure}

Bevel-tip steerable needles have the potential to provide minimally invasive access to anatomical sites deep in the human body~\cite{Abolhassani2007_MEP,Reed2011_RAM,Webster2006_IJRR,Rox2020_IA}.
These needles leverage asymmetric tip forces to curve around anatomical obstacles during needle insertion, enabling accurate targeting of clinically relevant sites that are difficult or impossible to reach safely with traditional needles.
The design trend has been to maximize the needle's curvature capability to increase reachability to many areas of the body in complex anatomy~\cite{vandeBerg2014_TMECH,Yang2018_JMRR}.
However, with an increase in curvature, the needle exerts more force on the surrounding tissue during deployment due to redirecting insertion forces~\cite{Reed2011_RAM}.
With large tissue forces perpendicular to the needle (see Fig.~\ref{fig:three-path-force}) comes an increased potential of significant tissue damage, such as tissue compression or a shearing event~\cite{Reed2011_RAM}.
A shearing event is where the needle shaft cuts sideways through the surrounding tissue, causing severe damage~\cite{Rox2020_IA} (see Fig.~\ref{fig:shearing}).

Minimum path length and maximum clearance are the most commonly used cost functions in needle steering~\cite{Patil2014_TRO,Kuntz2015_IROS,Fu2018_IROS,Pinzi2019_IJCARS}.
In the context of needle steering, these cost functions are mainly intended to encourage planned motions that minimize some notion of tissue damage by piercing through less total tissue (path-length cost) or steering far from highly sensitive anatomical structures (obstacle-clearance cost).

In this work we instead develop an efficient model describing the forces from the needle's shaft on surrounding tissue for a given path.
In our model, the force exerted by the needle on the surrounding tissue is a function of the puncture force at the needle's tip, the needle's shape through the tissue, and the friction between the needle's shaft and the surrounding tissue (see Fig.~\ref{fig:three-path-force}).

We develop a motion-planning cost function based on our force model.
We assume that the magnitude of the tissue normal force is correlated with the probability of tissue damage, from tissue compression to shearing.
We further assume that no shearing occurs below an unknown minimum force threshold.
Thus, we incorporate the maximum tissue force along the shaft as a cost function during motion planning, to effectively push down the force peak below the minimum force threshold.
This cost function, used in a suitable motion-planning algorithm, enables us to plan motions that reach clinically relevant targets while avoiding sensitive anatomical structures and directly minimizing the modeled normal force exerted upon the tissue by the needle shaft during insertion.
%

We propose this new clinically relevant needle force model and cost function as a replacement for currently used motion-planning cost functions with steerable needles.
Notably, with this model, the tissue normal forces are dependent on the entire needle trajectory and cannot be determined locally in isolation.
However, with a specified tip piercing force, we can compute the tissue normal forces in a single pass analytically, backward, starting at the needle's tip.
As a key result, our model shows that neither path length nor maximum curvature along a path can accurately serve as a proxy metric for the tissue normal forces along a path (see Fig.~\ref{fig:three-path-force}).

Our force-based cost function is an instance of a \emph{bottleneck cost}, which has the full cost concentrated in a localized piece of the trajectory.
In our case, the full cost is localized at the point of maximal tissue force.
We demonstrate the use of this force-based cost with a modified motion-planning algorithm.
This algorithm produces plans that achieve better costs, as computation time allows.

We provide the following contributions:
\begin{enumerate}
  \item
    a simple and efficient needle-to-tissue force and friction model;
  \item
    a new physically based and clinically relevant motion-planning cost function for needle steering based on our efficient needle-to-tissue force and friction model, and a motion-planning algorithm that leverages our force-based bottleneck cost function;
  \item
    two strategies for fitting our model parameters to experimental results without direct force measurements between the needle and tissue.
\end{enumerate}

We fit and validate our force model's ability to rank candidate paths with physical needle-steering experiments in gelatin, thus motivating its use as a motion-planning cost function.
We evaluate our motion planner with our force-based cost function by comparing against the same motion planner with the traditionally used path-length cost function in hundreds of 3-D randomly generated planning environments.
We compare the paths generated by the use of these two cost functions on both modeled normal force and path length.
Our motion-planning experiments show that, compared to using the path-length cost, planning with our force cost generates paths with a \SI{62}{\percent} decrease in maximal force at the cost of only \SI{0.07}{\percent} higher path length on average.
Finally, we demonstrate planning a force-cognizant motion plan in a lung tumor biopsy scenario segmented from a computed tomography (CT) scan.

By incorporating a cost function that explicitly models the interaction forces between the needle shaft and the surrounding tissue, motions can be planned for steerable needles that potentially reduce the risk of tissue damage with negligible impact on path length.
These contributions have the potential to reduce the risk of damage to sensitive anatomical structures and improve patient outcomes.

  \section{Related Work}
\label{sec:related-work}

Due to their potential to reduce the invasiveness of many types of therapeutic and biopsy-based procedures, steerable needles have been proposed for use in the kidneys~\cite{Majewicz2012_TBME}, liver~\cite{Majewicz2012_TBME}, prostate~\cite{Alterovitz2005_ICRA}, brain~\cite{Minhas2009_EMBS}, and lung~\cite{Swaney2017_JMRR}.
%
The needle-steering community has explored a wide range of needle actuation designs, including
beveled~\cite{Webster2006_IJRR},
pre-bent~\cite{Minhas2009_EMBS,Engh2006_EMBS},
passive flexure~\cite{Rox2020_IA,Swaney2012_TBME},
variable-length flexure~\cite{Bui2016_IME},
active flexure~\cite{Gerboni2017_RAL,vandeBerg2015_MEP},
fracture-directed inner stylet~\cite{Yang2018_JMRR},
a programmable bevel~\cite{Frasson2012_JRS},
and external magnetic actuation~\cite{Schwehr2022_HSMR,Ilami2020_NSR,Hong2020_TBME}.
See van de Burg \etal~\cite{vandeBerg2014_TMECH} for a review on steerable needle designs.
Most of these designs leverage an asymmetric tip, which causes them to curve in tissue as they are inserted from their base outside of the tissue~\cite{vandeBerg2014_TMECH,Rox2020_IA}.

Measuring and understanding the force interaction between the needle and tissue is important to minimizing tissue damage.
Force sensors have been placed on needle tips to more accurately measure the needle's piercing force and better understand the tip's interaction with different types of tissue and tissue boundaries~\cite{Gessert2019_IJCARS}.
High needle insertion force has been associated with excessive tissue damage~\cite{Gidde2020_BioinspirBiomim}.
Techniques have been used to decrease needle insertion force with barbs~\cite{Gidde2020_BioinspirBiomim}, vibrations~\cite{Gidde2020_BioinspirBiomim,Tsumura2019_JMRR}, bidirectional rotation~\cite{Tsumura2019_JMRR}, and slower insertion speeds~\cite{Webster2005_ICRA}.

Many have attempted to model the force interaction between the needle shaft and surrounding tissue, modeling insertion forces, tissue deformation, needle deflection, and cutting forces~\cite{Abolhassani2007_MEP,Oldfield2013_CMBBE,Takabi2017_MEP}.
These works primarily focus on finite-element simulations based on the full Cosserat-rod model and tissue mechanics~\cite{Antman2005_Chapter14}.
Although these finite-element simulations are fast enough to enable real-time control of steerable needles, they are not efficient enough for motion planning since optimal motion planning searches over the space of all possible valid trajectories, minimizing a cost metric.
Instead, our force model is based on the Cosserat-string formulation, which enables a simpler analytically tractable model that can be incorporated into existing motion planners.
This work is further differentiated by utilizing the force model as the cost function in motion planning with the goal of minimizing the tissue damage by needle shaft forces, including tissue compression and shearing.

Motion planning enables robots to plan trajectories that avoid obstacles while moving from some start state to a goal state.
Sampling-based motion planning is a popular paradigm that leverages random sampling of configurations or controls to produce collision-free motion~\cite{Salzman2019_CACM}.
These include the Rapidly exploring Random Tree (RRT)~\cite{LaValle2001_IJRR} and Probabilistic Roadmap (PRM)~\cite{Kavraki1996_TRA} methods which incrementally construct a collision-free tree or graph embedded in the configuration space.

Motion planning for steerable needles has been approached in a variety of ways.
Pinzi \etal~\cite{Pinzi2019_IJCARS} present the Adaptive Hermite Fractal Tree algorithm, which leverages optimized geometric Hermite curves~\cite{Yong2004_CAGD} combined with a so-called fractal tree.
Fu \etal~\cite{Fu2021_RSS} recently developed the Resolution-Complete Search (RCS) algorithm, and the resolution-optimal extension RCS*~\cite{Fu2022_ICRA}, which provably finds the lowest-cost needle-steering plan within the resolution of a discretized needle-steering action space.
Favaro \etal~\cite{Favaro2018_ICRA} adapt the Batch Informed Trees (BIT*) algorithm~\cite{Gammell2015_ICRA} combined with a path smoothing method in order to plan motions for a programmable bevel-tip needle.
Patil \etal~\cite{Patil2014_TRO,Patil2010_BioRob} built upon RRT to develop the Reachability-Guided RRT (RG-RRT) method for steerable needles.
RG-RRT has been adapted in other work to plan motions for a three-stage lung tumor biopsy robot~\cite{Kuntz2015_IROS}, and to plan in pulmonary cost maps automatically generated from medical imaging~\cite{Fu2018_IROS}.
We build upon RG-RRT in this work, with a modification that enables us to produce plans that achieve better costs, as computation time allows.

  \section{Method}
\label{sec:method}

We first derive our shaft-to-tissue force model, then we discuss our approach to incorporate this force model as a bottleneck cost function in a motion-planning context.

\subsection{Shaft-to-Tissue Force Model}
\label{subsec:force-model}
Here we derive a model for the forces exerted by a flexible needle shaft traveling on a planned path through tissue.
We model the needle as a Cosserat string~\cite{Antman2005_Chapter14} and incorporate a kinetic friction model to derive tissue forces as a function of the needle's planned path through the tissue.
The Cosserat string model assumes infinite flexibility.
This is a first-order approximation that does not account for all physical effects, but it makes the problem tractable.
Further, the assumption becomes more accurate as the bending stiffness of the needle decreases, relative to the tissue modulus, and is particularly relevant in light of recent advances in needle designs with decreased stiffness~\cite{Rox2020_IA,Yang2018_JMRR}.

Our derivation begins by decomposing the needle's forces into normal and tangential components.
We then incorporate a kinetic friction model that couples them together.
Integration of the governing equations along the shaft from the tip to the base then yields the predicted force distributions associated with the path.

\subsubsection{Model Assumptions}
As needles become thinner and more flexible, the bending stiffness vanishes, and the forces required to keep the needle in a static curved shape become negligible.
As the needle is pushed through tissue, the piercing force $\vec{F}_{\!p}$ must be transmitted from an insertion force $\vec{F}_{\text{ins}}$ at the base of the needle along the shaft until it reaches the needle's tip.
%
We assume that kinetic friction is the dominant force along the shaft and comes from the combination of compression from surrounding tissue and the needle pushing against the tissue along path curves.
The friction and normal forces are coupled in a way similar to the well-known capstan equation~\cite{Attaway1999_ITRS}.
Therefore, we model the needle inside the tissue as an ideal Cosserat string, which assumes that (i)~the flexural rigidity is negligible, and (ii)~the internal force vector is always tangent to the needle's path in space~\cite{Rucker2011_TRO,Antman2005_Chapter14}.

Conventionally, a Cosserat string is assumed to only carry tension force (since an ideal string will buckle under any compressive force).
However, we assume that compressive force can be carried without buckling because the surrounding tissue will constrain the needle and prevent buckling, even for very low stiffness needles.
The presented model is otherwise identical to a classical Cosserat string.

\subsubsection{Needle Equilibrium}
In this model, the needle is characterized by its centerline curve in space $\vec{p}(s) \in \mathds{R}^3$ as a function of the parameter $s \in [\num{0}, L]$.
In the following derivation, we use $s$ as the arc length along the needle path of length $L$.
The derivative of $\vec{p}(s)$ with respect to $s$, denoted as $\dot{\vec{p}}(s)$, is a unit vector tangent to $\vec{p}(s)$ pointing toward the robot's tip.

\begin{figure}[tpb]
  \centering
  {
    \def\svgwidth{\columnwidth}
    \Large
    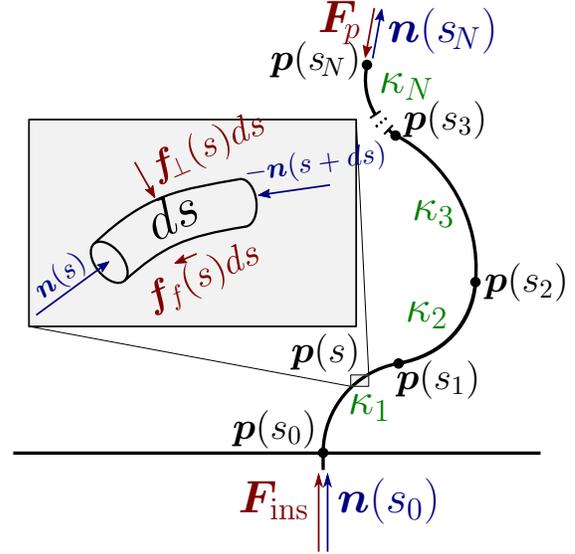
  }
  \caption[%
    An example piecewise-circular arc path with parameter endpoints $s_i$ and curvatures $\kappa_i$.
  ]{%
    An example piecewise-circular arc path with parameter endpoints $s_i$ and curvatures $\kappa_i$.
    A small segment of length $ds$ is shown with forces.
    Friction~$\vec{f}_f(s)$, tissue normal force $\vec{f}_{\!\bot}(s)$, internal needle force $\vec{n}(s)$, insertion force $\vec{F}_{\text{ins}}$, and piercing force $\vec{F}_{\!p}$ are labeled.
    }
  \label{fig:needle-path-and-piece}
\end{figure}

Along the needle's path, the tissue exerts a distributed force on the needle shaft which can be decomposed into two components, as seen in Fig.~\ref{fig:needle-path-and-piece}: one parallel to the needle, $\vec{f}_f(s) = -f_f(s) \dot{\vec{p}}(s)$, representing friction; and one perpendicular to the needle, $\vec{f}_{\!\bot}(s)$, representing the net normal force from tissue.

The force balance on a small needle section of length $ds$ is then
\begin{equation*}
  - \vec{n}(s + ds)
  +
  \vec{n}(s)
  +
  \vec{f}_{\!\bot}(s) ds
  -
  f_f(s) \dot{\vec{p}}(s) ds
    =
      \num{0},
\end{equation*}
where $\vec{n}(s)$ is the internal force vector carried by the needle, parallel to the needle path toward the needle's tip, and representing the transfer of $\vec{F}_{\mathrm{ins}}$ to $\vec{F}_{\!p}$.
The inset in Fig.~\ref{fig:needle-path-and-piece} shows the $\vec{n}(s)$ vector is the force that the proximal material exerts on the distal material.
%
%
Dividing by $ds$ and allowing $ds \rightarrow \num{0}$, then
\begin{equation} \label{eqn:force-equation}
  -\dot{\vec{n}}(s)
  +
  \vec{f}_{\!\bot}(s)
  -
  f_f(s) \dot{\vec{p}}(s)
    =
      \num{0},
\end{equation}
where the dot represents the derivative with respect to $s$.
This is the conventional Cosserat-string equilibrium equation~\cite{Rucker2011_TRO,Antman2005_Chapter14}, with the distributed force separated into two orthogonal components.
The assumption that the internal force vector $\vec{n}(s)$ is aligned with the tangent vector $\dot{\vec{p}}(s)$ implies that the needle cannot carry internal shear loads.
Thus
\begin{align*}
  \vec{n}(s)
    &=
      n(s) \dot{\vec{p}}(s)
  \\
  \dot{\vec{n}}(s)
    &=
      \dot{n}(s) \dot{\vec{p}}(s)
      +
      n(s) \ddot{\vec{p}}(s),
\end{align*}
where the scalar $n(s) = \|\vec{n}(s)\|$ represents the compressive force carried by the needle shaft at $s$.
Substituting these into~\eqref{eqn:force-equation} and decomposing into the parallel and perpendicular components, we get
\begin{align}
  \dot{n}(s)
    &=
      -f_f(s)
    \label{eqn:ndot}
  \\
  \vec{f}_{\!\bot}(s)
    &=
      n(s) \ddot{\vec{p}}(s).
    \nonumber
\end{align}
For any path-length parameterized curve $\vec{p}(s)$, the magnitude of $\ddot{\vec{p}}(s)$ is the curvature, $\kappa(s)$, thus
\begin{equation}  \label{eqn:ft-as-n}
  f_{\!\bot}(s)
    =
      \kappa(s)
      n(s),
\end{equation}
where $f_{\!\bot}(s)$ is the magnitude of $\vec{f}_{\!\bot}(s)$.

\subsubsection{Friction Model and Force Integration}
To calculate $n(s)$, the magnitude at one point must be given.
Typical points are either at the beginning (the insertion force $n(\num{0}) = F_{\text{ins}}$) or the end (the piercing force $n(L) = F_{\!p}$) as depicted in Fig.~\ref{fig:needle-path-and-piece}.
For a given needle path, we consider a known piercing force magnitude, $F_{\!p}$.
We assume the magnitude of the needle's insertion force, $F_{\mathrm{ins}}$, is sufficiently large to overcome friction and provide the needed piercing force, $F_{\!p}$, at the tip.
However, it could be easily adapted to a known insertion force magnitude, $F_{\text{ins}}$, if that is measured or controlled, thus using the model to predict the current piercing force magnitude, $F_{\!p}$.
In our evaluation, we assume a constant insertion speed and piercing force $F_{\!p}$.
We also note that the magnitude of the piercing force (and that of the resulting force distributions along the needle) need not be highly accurate in order to be informative for distinguishing between higher- and lower-force paths for planning purposes.

We assume a kinetic friction model for $f_f(s)$ of the form
\begin{equation} \label{eqn:friction-model}
  f_{f}(s)
    =
      \mu(s) \big(f_c(s) + f_{\!\bot}(s)\big),
\end{equation}
where $\mu(s)$ is the conventional coefficient of kinetic friction, $f_{\!\bot}(s)$ is the needle shaft's normal force on the tissue, and $f_c(s)$ is the distributed compressive force of the surrounding squeezing tissue; $\mu(s) f_c(s)$ is the resulting nominal distributed frictional force that would be present even for a straight needle path.
Substituting this friction model into~\eqref{eqn:ndot}, we arrive at the following first order linear differential equation
\begin{equation}
  \dot{n}(s)
    =
      - \mu(s) f_c(s)
      -
      \mu(s)
      \kappa(s)
      n(s).
      \label{eqn:n_ode}
\end{equation}
Solving this, subject to an initial or final condition, yields the internal compression force in the needle, from which the tissue normal force can be calculated via~\eqref{eqn:ft-as-n}.
In general, $f_c(s)$ and $\mu(s)$ could vary along $s$ as the needle passes through heterogeneous tissues;
$F_{\!p}$ may vary for each intermediate needle shape through heterogeneous tissue or tissue state.
Additionally, $\mu(s)$ and $F_{\!p}$ may depend on the rotational velocity of the needle~\cite{Tsumura2019_JMRR}.
We can numerically integrate~\eqref{eqn:n_ode} backward from the tip to the base starting with $n(L) = F_{\!p}$, and substitute the solution into~\eqref{eqn:ft-as-n} to calculate the tissue normal force distribution along the needle's path.
Alternatively, we can also express the general solution for $n(s)$ as
\begin{align*}
  n(s)
    &=
      A
      e^{-B(s)}
      -
      e^{-B(s)}
      \int
      \mu(s) f_c(s)
      e^{B(s)}
      ds
  \\
  B(s)
    &=
      \int \mu(s) \kappa(s) ds,
\end{align*}
where $A$ is a constant of integration that can be determined by applying the tip condition $n(L) = F_{\!p}$, and where, depending on the nature of the functions $f_c(s)$, $\mu(s)$, and~$\kappa(s)$, the integrals can either be evaluated analytically or numerically.

\begin{figure}[tb]
  \centering
  {
    \def\svgwidth{\columnwidth}
    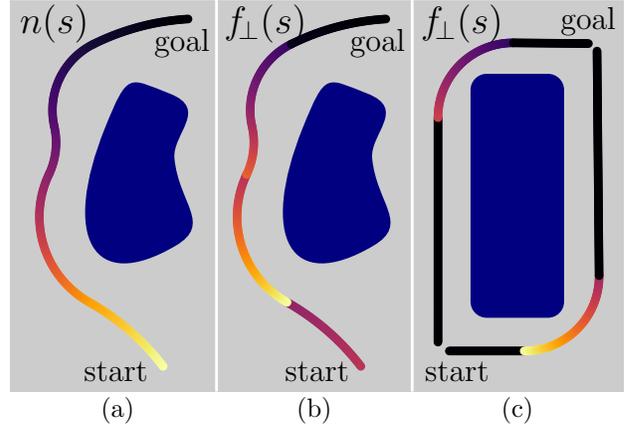
  }
  \caption[%
    Force heat maps along the needle path using this work's new tissue normal force model.
  ]{%
    Force heat maps along the needle path using this work's new tissue normal force model.
    Darker colors indicate small values, lighter colors indicate large values.
    \textbf{(a)}~Internal compression force $n(s)$ carried by the needle from insertion to tip piercing force.
    \textbf{(b)}~Resulting magnitude of the normal force exerted from the needle on the tissue $f_{\!\bot}(s)$ for the same path followed in (a).
    \textbf{(c)} Two paths from start to goal of the same length and of the same two straight segments and one arced segment, which demonstrates that high curvature at the beginning of the path results in much higher tissue normal forces than high curvature near the end.
    %
    %
    In (c), the maximum tissue normal force of the lower-right path is \SI{74}{\percent} higher than the maximum tissue normal force of the upper-left path. 
  }
  \label{fig:tissue-force-heatmap}
\end{figure}

If $f_c(s)$, $\mu(s)$, and $\kappa(s)$ are piecewise constant---say $f_{c,i}$, $\mu_i$, and $\kappa_i$ on $s \in (s_{i-1}, s_i)$ as in Fig.~\ref{fig:needle-path-and-piece}, and let $n_i(s) = n(s)$ and $f_{\!\bot,i}(s) = f_{\!\bot}(s)$ for the $i^{\text{th}}$ segment---then for $s \in (s_{i-1}, s_i)$, and $\kappa_i > 0$, the solution reduces to
\begin{equation} \label{eqn:n-solved}
  n_i(s)
    =
      -
      \dfrac{f_{c,i}}{\kappa_i}
      +
      \left(
        n_{i+1}(s_i) + \dfrac{f_{c,i}}{\kappa_i}
      \right)
      e^{
        \mu_i \kappa_i (s_i - s)
      },
\end{equation}
with~\eqref{eqn:ft-as-n} becoming
\begin{equation*}
  f_{\!\bot,i}(s)
    =
      \kappa_i n(s).
\end{equation*}
For zero-curvature sections, $\kappa_i = 0$, $f_{\!\bot,i}(s) = 0$, and $n_i(s) = n_{i+1}(s) + \mu_i f_{c,i} (s_i - s)$.
This solution can be iteratively evaluated section by section, starting at $s_i = L$ and proceeding backward to the base.
Note that the $n(s)$ solution is continuous across the entire needle trajectory, whereas $f_{\!\bot}(s)$ is discontinuous due to possible curvature discontinuities at each $s_i$, as illustrated in Fig.~\ref{fig:tissue-force-heatmap}(a) and~\ref{fig:tissue-force-heatmap}(b) respectively.

If we choose to ignore the effects of friction (by setting $\mu(s) = \num{0}$), we get $f_{\!\bot}(s) = \kappa(s) F_{\!p}$, which results in a tissue normal force directly proportional to the curvature.
Thus, while it may be intuitive to assume the curvature itself would be a good proxy for the probability of tissue compression or shearing, our model predicts that assumption is only true in the absence of shaft-to-tissue friction.
If we ignore the influence of $f_{\!\bot}(s)$ on friction (by removing $f_{\!\bot}(s)$ from~\eqref{eqn:friction-model}, assuming $f_{\!\bot}(s) \ll f_c(s)$), we get a simple linear internal force model for~\eqref{eqn:n_ode} with friction being independent of needle shape.
If we consider the full friction model, even for a small friction coefficient~$\mu_i$, the required insertion force for a given piercing force grows exponentially with the product of the friction coefficient, length, and curvature.
As the path gets longer, the internal needle force, $n(s)$, grows exponentially as seen in~\eqref{eqn:n_ode}, thus predicting a sharp increase in the tissue-damage probability.

\subsubsection{Friction Models}
In this work, we consider friction models that consist of constant $f_c$, $\mu$, and $F_{\!p}$ over the full needle shape.

The first friction model we consider ignores the relation of the needle rotational velocity on friction behavior, and uses the same $f_c$, $\mu$, and $F_{\!p}$ for the full insertion trajectory.
%
We call this the \emph{single-parameter} model in our evaluation.

Because of the observation that insertion force decreases with needle rotation~\cite{Tsumura2019_JMRR}, we consider a second friction model that utilizes separate $\mu$ and $F_{\!p}$ parameters for when the needle is rotating or non-rotating.
We assume the compressive tissue force $f_c$ is unaffected by needle rotation, and therefore, this second friction model uses a single $f_c$ regardless of the needle rotational velocity.
We call this the \emph{double-parameter} model in our evaluation.

To clarify, when the needle is rotating, the needle tip moves in a straight path.
When the needle is non-rotating, the needle tip moves in a constant curvature arc in the direction of the bevel tip, specifically at the maximum curvature $\kappa_{\text{max}}$.
To achieve curvature between \num{0} and $\kappa_{\text{max}}$, we employ duty cycling between rotating and non-rotating states proportional to the desired percentage of $\kappa_{\text{max}}$ curvature~\cite{Minhas2007_EMBS,Minhas2009_EMBS,Qi2022_TIMC}.
Although rotating and non-rotating sections dictate the needle tip's path, when the needle is rotating, the rotating friction coefficient, $\mu_{\mathrm{rot}}$, applies to the entire path; likewise for the non-rotating friction coefficient, $\mu_{\mathrm{norot}}$.

\subsubsection{Model Fitting}
Measuring the normal forces between the needle and tissue is difficult due to the challenge of instrumenting such thin tissue-embedded needles without interfering with the force interaction.
However, a key insight of our work is that the model can be fit without such measurements.
We present two fitting methods.
%
The first method fits based on the axial insertion forces, which can be measured by instrumenting the needle outside of the tissue.
However, this is itself also difficult as it is non-trivial to completely capture all forces at the needle insertion site while rotating the needle and constraining it from buckling in air.
The second, and potentially less burdensome method, is to fit the model using a set of labeled shearing events.

When using measured insertion forces, $F_{\text{ins}}$, one may fit the model parameters using any non-linear least squares solver, fitting on the error in $F_{\mathrm{ins}}$ between measured values and the model's prediction.
Alternatively, if given an average shearing force $\bar{f}_{\mathrm{shear}}$, our model can predict the depth of shearing by determining the earliest point during path execution that exceeds $\bar{f}_{\mathrm{shear}}$ over a given trajectory.
%
This strategy is the basis of our second fitting method; one may similarly fit the model parameters using any non-linear least squares solver, fitting on the error between the predicted and measured shearing depths.

For models fit against measured $F_{\mathrm{ins}}$, the average shear force $\bar{f}_{\mathrm{shear}}$ can be viewed as an additional parameter of the model, and can be fit to labeled shearing depths.
We simply use the dataset's sample mean of the model's predicted maximal $f_{\!\bot}(s)$ over the needle's shapes at the labeled shearing depths.

Given that we use these fit models exclusively as cost functions in motion-planning, the exact magnitude of predicted needle-tissue force need not be fully accurate.
What matters is the model's ability to rank paths relative to each other, with the goal of generating trajectories with all $f_{\!\bot}(s)$ below the unknown minimum shearing threshold, $f_{\textrm{thresh}}$.
To this end, we evaluate our models by comparing the ranking they assign to paths against the true ranking (from labeled shearing depths).
Since path ranking is scale-invariant, the $\bar{f}_{\text{shear}}$ used in fitting against measured shearing depths is functionally arbitrary.

We use the Trust Region Reflective algorithm~\cite{Branch1999_JSC} for non-linear least squares.
We constrain all parameters to be strictly positive to represent meaningful physical quantities.
For the double-parameter model, we further constrain $F_{p,\text{rot}} \leq F_{p,\text{norot}}$ and $\mu_{\text{rot}} \leq \mu_{\text{norot}}$ to match the prior observation of lower insertion force during needle rotation~\cite{Tsumura2019_JMRR}.


\subsection{Motion Planning}
We propose a motion-planning framework that enables planning of needle trajectories with a lower risk of tissue damage by minimizing the modeled normal forces being applied by the needle to the tissue during insertion, thus increasing the likelihood that $f_{\!\bot}(s) < f_{\mathrm{thresh}}$ for all $s$.

\subsubsection{Bottleneck Cost} 

For a needle trajectory $\vec{\pi}$ parameterized by time, $t \in [0, t_f]$, with $t_f$ representing the final time of trajectory $\vec{\pi}$, respectively, let $\vec{p}_t(s)$ be the needle's centerline curve, as defined in Section~\ref{subsec:force-model}, at time $t$ of trajectory $\vec{\pi}$ (denoted $\vec{\pi}(t)$); likewise let the $t$ subscript refer to the associated function over the needle shape at $\vec{\pi}(t)$.
We define our force-based cost as the maximal tissue normal force along any needle shape within the trajectory,
\begin{equation*} 
  C_F(\vec{\pi})
    =
      \max_{t \in [0, t_f]}
      \max_{s \in [0, L_t]}
      f_{\!\bot,t}(s),
\end{equation*}
with $f_{\!\bot,t}(s)$ determined by our force model above in~\eqref{eqn:ft-as-n}.

This $C_F(\vec{\pi})$ cost function is an example of a bottleneck cost, a concept that has been extensively studied in the motion-planning community (see, e.g., from Solovey \etal~\cite{Solovey2017_IROS} and references within) with diverse applications such as following manipulator and surgical trajectories~\cite{Holladay2019_RAL, Niyaz2019_IROS}.


Formally, if $c$ is a point-wise cost along a trajectory $\vec{\pi}$, then we can construct its associated \emph{bottleneck cost} as
\begin{equation*}
  C(\vec{\pi}) = \max_{t \in [0, t_f]} c(\vec{\pi}(t)).
\end{equation*}


Our point-wise trajectory cost $c_F$ (also referred to as the needle shape cost) is the maximal tissue normal force along the needle shape at time $t$,
\begin{equation*}
  c_F(\vec{\pi}(t))
    =
      \max_{s \in [0, L_t]}
      f_{\!\bot,t}(s).
\end{equation*}
We use the common follow-the-leader assumption which states that the needle shape at $\vec{\pi}(t)$ is equal to the needle tip trajectory from 0 to $t$; i.e., if $t < t'$ and $s \in [0, L_t]$, then $\vec{p}_{t}(s) = \vec{p}_{t'}(s)$.
In the case of constant model parameters during the trajectory, the maximal shape cost is at the trajectory's end, $\vec{\pi}(t_f)$.
However, even in that case, our cost $C_F$ maximizes over the needle shape and still resembles a bottleneck cost.

One key property of bottleneck costs is that a trajectory's cost is the max of its subtrajectories' costs, i.e., if $t_1 < t_2 < t_3$, then
\begin{equation*}
  C(\vec{\pi}[t_1, t_3]) = \max(C(\vec{\pi}[t_1, t_2]), C(\vec{\pi}[t_2, t_3]));
\end{equation*}
whereas accumulation-based costs are additive,
\begin{equation*}
  C(\vec{\pi}[t_1, t_3]) = C(\vec{\pi}[t_1, t_2]) + C(\vec{\pi}[t_2, t_3]).
\end{equation*}

%
%
Existing optimizing motion planners either require (i)~solving the two-point boundary value problem (see, e.g.,~\cite{Karaman2011_IJRR,Salzman2016_TRO,Solovey2017_IROS}) or (ii)~cannot be easily adapted to use a bottleneck cost.

\subsubsection{Approach}
As we are not aware of any existing method to efficiently solve the two-point boundary value problem for steerable needles, we introduce a general simple-yet-effective framework.
%
Our approach, summarized in Alg.~\ref{alg:planner}, takes as input any sampling-based motion planner \textit{ALG} that can efficiently discard candidate paths below a given cost threshold $c_{\text{max}}$ during its search, such as discarding edges that cause the path's cost to exceed $c_{\text{max}}$.\footnote{
  This is a very natural assumption;
  any planner that maintains paths using a configuration-space graph can typically be adapted to discard candidate paths below a given cost threshold.
}
Our algorithm also requires an approximation factor $\varepsilon$, which controls how aggressively the cost threshold geometrically decreases after each successive found candidate trajectory.
%
%
For the first solution to be unconstrained, we initialize $c_{\text{max}}$ to infinity (Alg.~\ref{alg:planner} line~\ref{alg:planner:line:init-cmax}).
Once a solution is obtained, the maximal cost value $c_\textrm{max}$ is updated to be $C(\vec{\pi})/(\num{1}+\varepsilon)$ (Alg.~\ref{alg:planner} line~\ref{alg:planner:line:set-cmax}), where $C$ is the cost function and $\vec{\pi}$ is the solution returned by \textit{ALG} (Alg.~\ref{alg:planner} line~\ref{alg:planner:line:call-alg}).

\begin{algorithm}[t]
  \caption{Bottleneck-Cost Planner}
  \label{alg:planner}
  \begin{algorithmic}[1]
    \Procedure{Planner}{\textit{ALG}, $\vec{q}_0$, $G$, $O$, $C$, $\varepsilon$}
    \State{\textit{ALG}: sampling-based motion planner}
    \State{$\vec{q}_0$: start configuration}
    \State{$G$: goal configuration set}
    \State{$O$: obstacle set}
    \State{$C$: cost function}
    \State{$\varepsilon$: approximation parameter}
      \State{$c_\textrm{max} \leftarrow \infty$}
        \label{alg:planner:line:init-cmax}
      \While{\emph{time allows}}
        \State{$\vec{\pi}
          \leftarrow
            \textit{ALG}(\vec{q}_0, G, O, C, c_\textrm{max})$ }
          \label{alg:planner:line:call-alg}
        \State{$c_\textrm{max} \leftarrow C(\vec{\pi}) /(1 + \varepsilon)$}
          \label{alg:planner:line:set-cmax}
        \State{\textbf{report} $\vec{\pi}$}
          \label{alg:planner:line:return-plan}
      \EndWhile
    \EndProcedure
  \end{algorithmic}
\end{algorithm}

If $c^*$ is the optimal cost (i.e., $c^* = \inf_{\vec{\pi}} C(\vec{\pi})$), then no progress can be made if $c_{\text{max}} < c^*$, in which case, the previously returned plan would have cost $c = (1 + \varepsilon) c_{\text{max}} < (1 + \varepsilon) c^*$.
A large $\varepsilon$ may reach this point of no progress sooner than a small $\varepsilon$, but may ultimately find a final trajectory with higher cost.

It is worth noting that our framework bares resemblance to recent approaches~\cite{Hauser2016_TRO,Kleinbort2020_ICRA} to compute an asymptotically optimal path when path cost is additive (and not a bottleneck cost).


To apply this proposed framework to our needle-steering domain, we use as \textit{ALG} the RG-RRT algorithm~\cite{Patil2010_BioRob}---a state-of-the-art motion planner for needle steering.
Roughly speaking, RG-RRT runs an RRT-like algorithm but it extends robot configurations in the search tree toward a workspace needle tip position (where needle orientation is unconstrained) and \emph{not} a randomly sampled configuration (that includes both the needle's position and orientation).
This property motivates our use of RG-RRT as extending the tree toward a workspace region (and not a configuration-space region) enables the planner to employ goal biasing; a well-established strategy for increasing the speed of motion planning.
%
%

\subsubsection{Implementation}
%
%
%
%
%
For our setting we make two changes to RG-RRT: (i) instead of planning from the needle-insertion site to the target in the body, we perform a backward search and plan from the target in the body to a specified insertion site, and (ii) if an edge under consideration would exceed the provided cost threshold $c_{\text{max}}$, then we mark it as invalid.

The backward search is due to our choice of assuming a known constant piercing force $F_{\!p}$, and the model thus solving for the forces backward, as in~\eqref{eqn:n-solved}.
%
Because of the follow-the-leader assumption we only need to consider the final needle shape.
If we have a piecewise-constant $F_{\!p}$, we would need to calculate the maximal force over all subshapes where the tip is at a piercing force discontinuity.
This computation could be done as a single backward pass over the final shape, accounting for each individual subshape ending at piercing force discontinuity boundaries.

For constant piercing force $F_{\!p}$,  we only need to consider the final shape when computing $f_{\!\bot}(s)$, which depends on $\kappa_i$ and $n(s)$ in~\eqref{eqn:ft-as-n}.
This is because a path extension can only increase $f_{\!\bot}(s)$, since $\kappa_i$ is a constant, but $n(s)$ grows exponentially.
To compute $f_{\!\bot}(s)$ along a single needle shape, we start at the tip where $n(L) = F_{\!p}$, compute $n(s)$ backward one segment at a time using~\eqref{eqn:n-solved}, and use~\eqref{eqn:ft-as-n} to compute $f_{\!\bot}(s)$.
We store the internal needle force $n(s)$ as part of the motion-planning state, which, combined with planning from the goal toward the start, enables fast computation of $f_{\!\bot}(s)$ for each added segment.
Each segment is checked against the maximum-allowed cost when added and pruned if that cost is exceeded (see $c_{\textrm{max}}$ in Alg.~\ref{alg:planner} line~\ref{alg:planner:line:call-alg}).
The resulting found plan from the goal to the start is then reversed to create a planned trajectory.
Planning the path as if it were reversed in direction enables efficient and accurate computation of the maximum normal forces during forward insertion.

When considering the double-parameter model, where we have separate $\mu$ and $F_{\!p}$ values for rotating and non-rotating needle states, our motion planner keeps track of two separate $n(s)$ values as it propagates from the needle tip to the base.
We track the rotating state from the first segment (from the goal) that exhibits needle rotation, and likewise for the non-rotating state.
A segment is considered rotating in our implementation if $\kappa \leq 0.9 \kappa_{\text{max}}$ and is non-rotating if $0.1 \kappa_{\text{max}} \leq \kappa$.
Note, that for $0.1 \leq \frac{\kappa}{\kappa_{\text{max}}} \leq 0.9$, it is considered both rotating and non-rotating because of duty cycling (switching between the two).
The final cost value is then the maximum of the two predicted tissue forces at any point along the path shape.

  %
%
%
%
%

\section{Experiments}
\label{sec:experiments}
We first validate our force model's ability to rank candidate paths through physical experiments.
We then use one set of fit parameters to compare the effectiveness of planning for needle steering with our force-based cost function (the force planner) compared to using the path-length cost (the length planner).
We perform planning in 400 randomly generated environments.
We compare the plans from the force and length planners by their path lengths and maximal tissue forces.
Finally, we demonstrate the use of our force planner within a simulated lung biopsy scenario, planning through a segmented lung CT.

\subsection{Physical Experiment}
We validate the use of our needle model in motion planning using physical needle experiments in gelatin phantoms.
In motion planning, the cost function provides scores to effectively define the best-to-worst ranking of candidate paths.
To evaluate our cost function's ranking ability, we use various paths through a gelatin phantom and generate a true ranking based on the observed shearing insertion depth.
We compare our fit models' path ranking, based on predicted shearing depths, against the true ranking, based on measured shearing depths.

\begin{figure}[tpb]
  \centering
  {
    \def\svgwidth{\columnwidth}
    \scriptsize
    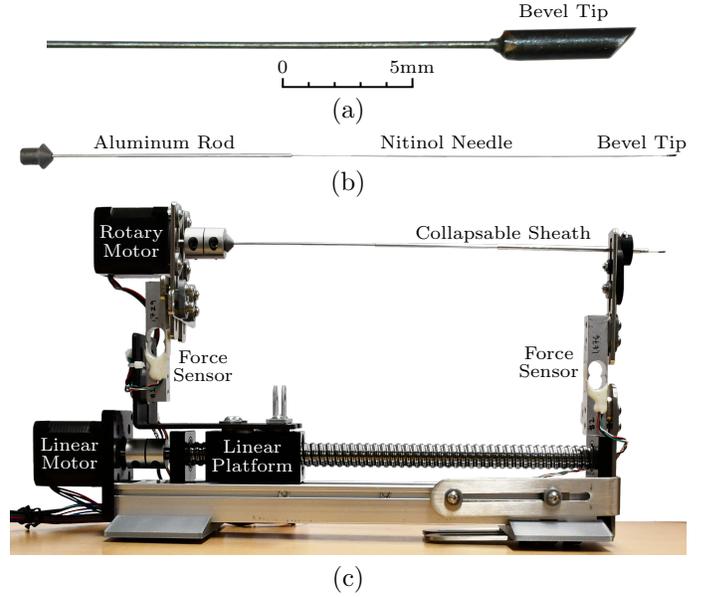
  }
  \caption[%
    The needle and robot design for our physical needle-steering experiments.
  ]{%
    The \textbf{(a)} needle bevel tip, \textbf{(b)} full needle, and \textbf{(c)} robot design for our physical experiments.
  }
  \label{fig:physical-robot}
\end{figure}

Our needle, shown in Fig.~\ref{fig:physical-robot}(b), consists of a nitinol tube with a large bevel tip (see Fig.~\ref{fig:physical-robot}(a)).
The thin \SI{165}{\mm} flexible nitinol tube section has a \SI{0.37}{\mm} outer diameter (OD) and a \SI{0.24}{\mm} inner diameter (ID).
The \SI{5}{\mm} bevel tip is \SI{1.22}{\mm} OD, \SI{1.02}{\mm} ID, filled with cyanoacrylate, and beveled to approximately \SI{45}{\degree}.
For stability and extra length to account for the collapsed sheath, the thin nitinol tube is affixed to a \SI{107}{\mm} aluminum rod with \SI{1.59}{\mm} OD.
Note that the aluminum rod remains outside of the tissue during all insertions.

The robot, shown in Fig.~\ref{fig:physical-robot}(c), is actuated by two Nema 17 stepper motors, one for linear actuation on a \SI{5}{\mm} pitch lead screw, and the other for rotary actuation of the needle itself.
The rotary motor connects directly to the steerable needle, which goes through a two-segment collapsible aluminum sheath to prevent the needle from buckling in air and at the entry point~\cite{Webster2005_ICRA}.
Both the sheath and rotary motor are affixed to the robot base through two \SI{1}{\kg} 1-D strain-gauge force sensors measured at \SI{11}{\Hz}.
We add both force sensor signals together to measure total axial force from the needle.


We push the needle through a gelatin phantom comprised of \SI{10}{\percent} Knox gelatin powder and \SI{90}{\percent} water, by weight.
%
%
Above the gelatin, a camera records insertions for later labeling of curvature and shearing (see Fig.~\ref{fig:shearing}).

We insert the needle at a constant insertion speed of \SI{5}{\mm\per\s}.
When the needle is rotating, we use a constant rotational velocity of \SI{2}{\rev\per\s}.
We achieve curvature less than $\kappa_{\mathrm{max}}$ with duty cycling between zero and \SI{2}{\rev\per\s}.
Our setup does not provide position feedback, so we use open-loop control for needle steering.

\subsubsection{Results}

%
%
We gathered 35 individual runs on a wide variety of controlled paths.
The full insertion length of each path is \SI{150}{\mm}, or until we observe shearing, in which case insertion is stopped early.
All paths start with a straight segment (i.e., the needle rotating) from between \SI{10}{\mm} to \SI{100}{\mm}.
Of the 35 paths, 19 of them contained at least one duty-cycled segment from \SIrange{70}{95}{\percent}.
Before beginning a run, the needle is inserted \SIrange{1}{2}{\mm} manually and adjusted until the force sensor readings are nearly at zero.

To perform a fit and generate predictions from our model, we require the path curvature at each point.
On 13 paths, we labeled the change in orientation $\Delta\theta$ from video footage and controlled insertion length $\Delta s$, which results in $\kappa = \sfrac{\Delta\theta}{\Delta s}$.
We estimated an average $\kappa_{\mathrm{max}} = \SI{0.020}{\mm^{-1}}$ and standard deviation \SI{0.003}{\mm^{-1}} (turning radius of \SI{50}{\mm}).
We use $\kappa_{\mathrm{max}} = \SI{0.02}{\mm^{-1}}$ in all fitting and planning in this paper.
%

For brevity, we refer to the single- and double-parameter models fit from measured $F_{\mathrm{ins}}$ as \ModelForceSingle and \ModelForceDouble, respectively.
We likewise define \ModelShearSingle and \ModelShearDouble as the models fit from labeled shearing depths.
When fitting against shear depth, we use a constant $\bar{f}_{\mathrm{shear}} = \SI{17.6}{\mN\per\mm}$.
This value was chosen to be near to fits generated from the insertion force.
We chose this value to enable the two models to output values on similar scales.
As the ranking is scale-invariant, this has no impact on the models' ability to rank paths.

%
%
We fit our four models from a randomly selected dataset of size \num{12} with \num{23} held out (for a total of \num{35} runs).
These parameters are seen in Table~\ref{tbl:models}.
We note that the parameters vary quite heavily between the models.
The \ModelShearSingle and \ModelShearDouble models are only expected to accurately predict shearing depth, not specific force values; therefore, their fit parameters may exhibit large variance.
The primary difference between Shear and Force model parameters is the predicted friction coefficient, with the force-fit models predicting very small friction and the shear-fit models predicting very high friction.
Of note, the 10 different fits of 12 runs varied widely in fit parameters, but their ability to accurately rank paths was not drastically impacted.
For example, the $\mu_{\mathrm{rot}}$ parameter had a median fit (with lower and upper quartiles specified with ``Q:'') of
\numquart{0.019}{0.0001}{0.13},
\numquart{0.0003}{0.0001}{0.12},
\numquart{0.30}{0.04}{0.61}, and
\numquart{0.33}{0.30}{0.36}
for
\ModelForceSingle,
\ModelForceDouble,
\ModelShearSingle, and
\ModelShearDouble, respectively,
yet their $\tau_D$ correlations were
\numquart{0.871}{0.870}{0.888},
\numquart{0.889}{0.876}{0.896},
\numquart{0.905}{0.902}{0.908}, and
\numquart{0.920}{0.907}{0.927}, respectively over all 35 runs.

\begin{table*}[tbp]
  \centering
  \caption{
    Fit parameters and $\tau_D$ correlation against 12 runs for our two models and two fitting strategies.
  }
  \label{tbl:models}
  \begin{tabular}{@{}lccccccc|cc@{}}
    \toprule
    {\narrowdoublestack{Model}{~}}
      & {\narrowdoublestack{$F_{p,\mathrm{rot}}$}{(\si{\milli\N})}}
      & {\narrowdoublestack{$F_{p,\mathrm{norot}}$}{(\si{\milli\N})}}
      & {\narrowdoublestack{$\mu_{\mathrm{rot}}$}{~}}
      & {\narrowdoublestack{$\mu_{\mathrm{norot}}$}{~}}
      & {\narrowdoublestack{$\mu_{\mathrm{rot}} f_c$}{(\si{\milli\N\per\mm})}}
      & {\narrowdoublestack{$\mu_{\mathrm{norot}} f_c$}{(\si{\milli\N\per\mm})}}
      & {\narrowdoublestack{$\bar{f}_{\mathrm{shear}}$}{(\si{\milli\N\per\mm})}}
      & {\narrowdoublestack{$\tau_D$}{~}}
      & {\narrowdoublestack{$p$-value}{~}}
    \\
    \midrule
    {\ModelForceSingle} & 122.0 & 122.0 & 0.0371   & 0.0371   &  9.48 &  9.48 & 13.2 & 0.87 & \num{5.9e-7} \\
    {\ModelForceDouble} & 123.5 & 123.5 & 0.000258 & 0.000276 &  9.27 &  9.98 & 13.1 & 0.89 & \num{3.6e-7} \\
    {\ModelShearSingle} & 438.5 & 438.5 & 0.287    & 0.287    &  4.99 &  4.99 & 17.6 & 0.90 & \num{2.3e-7} \\
    {\ModelShearDouble} &  82.8 & 171.4 & 0.232    & 0.248    & 11.07 & 11.83 & 17.6 & 0.94 & \num{0.7e-7} \\
    %
    %
    \bottomrule
  \end{tabular}
\end{table*}

%
%

%
%

\begin{figure*}[tb]
  \centering
  {
    \def\svgwidth{0.8\textwidth}
    \footnotesize
    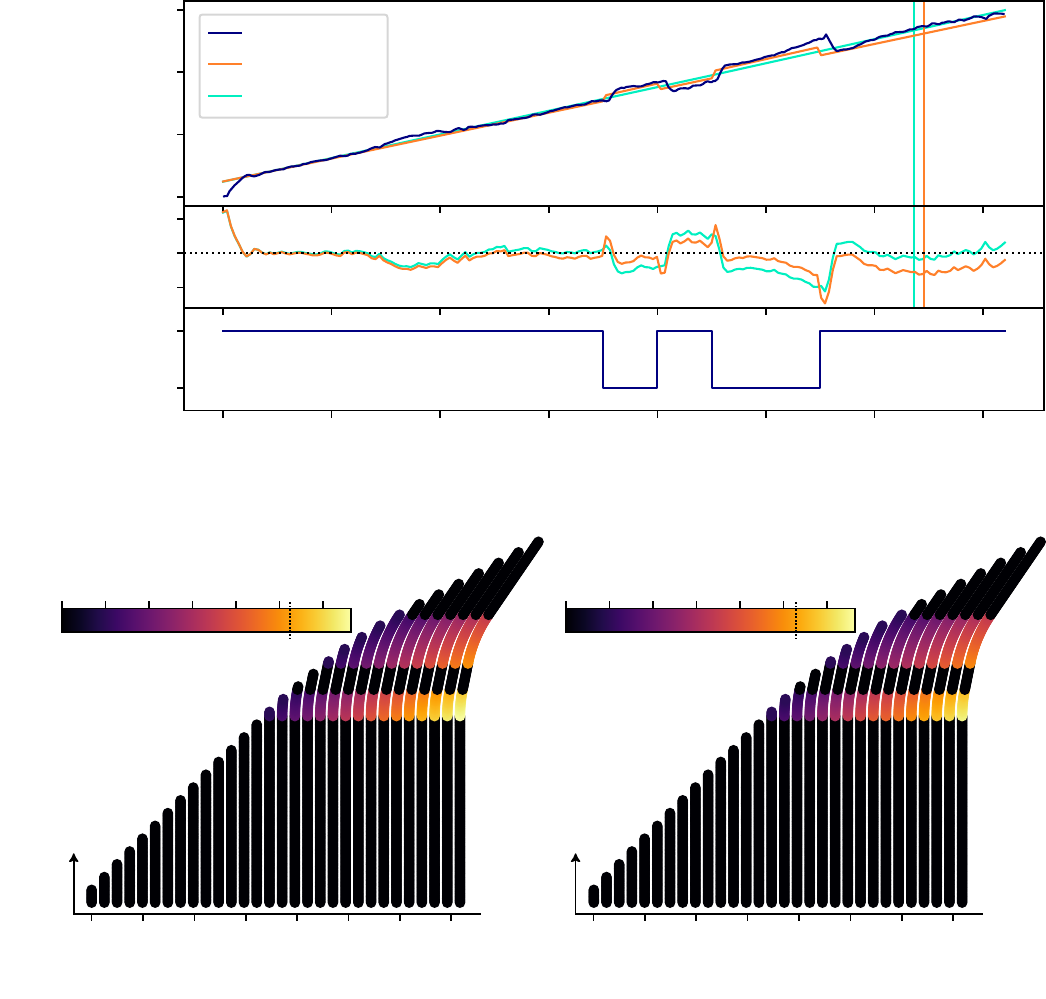
  }
  \caption[%
    Modeled and measured $F_{\mathrm{ins}}$ of a run used in fitting, and the models' predicted tissue force along the path.
  ]{%
    Modeled and measured $F_{\mathrm{ins}}$ of a run used in fitting, and the models' predicted tissue force along the path.
    \textbf{(a)} Plot of (top) measured and modeled $F_{\mathrm{ins}}$, (middle) modeling error in $F_{\mathrm{ins}}$, and (bottom) needle rotational velocity.
    The plot ends at the measured shearing, and the predicted shearing is indicated with vertical lines.
    The upward jumps in insertion force happen when the needle stops spinning, as seen in the rotational velocity subplot.
    \textbf{(b)} and \textbf{(c)} Intermediate needle shapes and modeled tissue force from \ModelForceSingle and \ModelForceDouble, respectively.
  }
  \label{fig:model-forces}
  \vspace{3em}
\end{figure*}

\begin{figure*}[tb]
  \centering
  {
    \def\svgwidth{0.83\textwidth}
    \footnotesize
    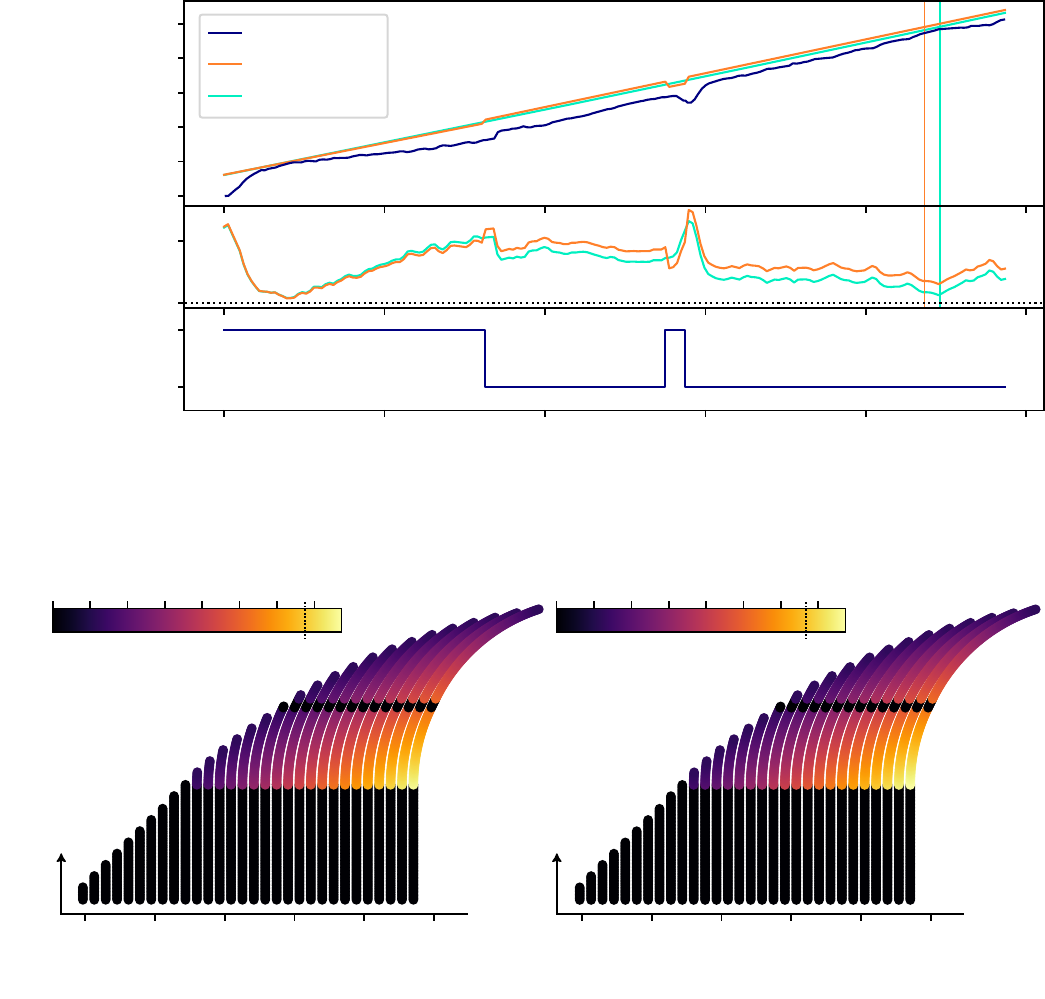
  }
  \caption[%
    Modeled and measured $F_{\mathrm{ins}}$ from a hold-out run, and the models' predicted tissue force along the path.
  ]{%
    Modeled and measured $F_{\mathrm{ins}}$ from a hold-out run, and the models' predicted tissue force along the path.
    This plot is structured the same as Fig.~\ref{fig:model-forces}, but shows a path not used in the fit.
    In this example, the predicted insertion forces from this particular fit do not fit the exponential behavior shown in the measured insertion force.
    Despite this insertion force mismatch, the predicted shearing depth for the \ModelForceDouble has nearly identical error percentage compared to Fig.~\ref{fig:model-forces} (\SI{10}{\percent}), and for the \ModelForceSingle, the error percentage is much better (\SI{12}{\percent} versus \SI{4}{\percent}).
  }
  \label{fig:model-forces-p30}
\end{figure*}

To demonstrate the fit quality and visualize our model's predicted forces, we display measured forces compared against predictions from the fit \ModelForceSingle and \ModelForceDouble models in Fig.~\ref{fig:model-forces}(a) and Fig.~\ref{fig:model-forces-p30}(a) (for one run from the fit dataset and the hold-out set, respectively).
%
%
%
The double-parameter model is able to account for jumps in the insertion force when the needle changes its rotational velocity.
However, the errors (middle plot), predicted tissue force profiles (Fig.~\ref{fig:model-forces}(b) and Fig.~\ref{fig:model-forces-p30}(b)), and predicted shearing depths (vertical lines in plots) of both models are very similar.
Although the force fit is worse against the hold-out run than the in-fit run, the final predicted shearing depth is better in this particular hold-out example.

%
%
To evaluate our model's ability to rank paths, we order the paths by shearing depth, comparing manually labeled shearing depths against the model predictions.
We use the similarity-weighted Kendall~$\uptau$ distance~\cite{Kumar2010_WWW}, a generalization of Kendall's~$\uptau$ weighted by element distances,
\begin{equation*}
  K_D = \sum_{i<j} D_{ij} \sigma_{ij},
\end{equation*}
where $\sigma_{ij}$ is one if elements $i$ and $j$ are out of order and $D_{ij}$ is their distance; i.e., the absolute difference between the actual shear depths.
The Kendall correlation coefficient is obtained from the Kendall distance by normalizing, inverting, and scaling to be between -1 and 1,
\begin{equation*}
  \uptau_D = 2 - \dfrac{K_D}{\displaystyle\sum_{i<j} D_{ij}}.
\end{equation*}
To obtain a $p$-value, we use the law of large numbers and a null hypothesis of random ordering.

To compute $\tau_D$, we evaluate the path ranking among all 35 paths, combining the fitting data and hold-out data.
For our fit models in Table~\ref{tbl:models}, we note that while the models have relatively different parameters, they nonetheless exhibit similar abilities to rank candidate paths with correlation coefficients between \numrange{0.87}{0.94} and $p$-values between \numrange{0.7e-7}{5.9e-7}.

\begin{figure}[tpb]
  \centering
  {
    \def\svgwidth{\columnwidth}
    \footnotesize
    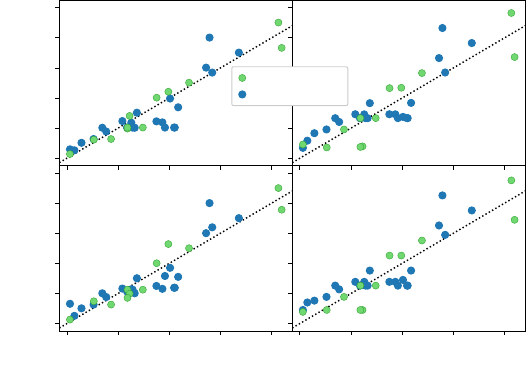
  }
  \caption[%
    Qualitative correlation visualization of measured shear depth versus predicted shear depth for various fits.
  ]{%
    Qualitative correlation visualization of measured shear depth versus predicted shear depth for various fits.
    %
    The model parameters are fit from the same set of \num{12} randomly selected runs, shown in light green, with the remaining \num{23} runs shown in blue.
    A perfect model would have all points on the $y=x$ diagonal dotted line.
    %
    %
    %
  }
  \vspace{-1em}
  \label{fig:correlation}
\end{figure}

In Fig.~\ref{fig:correlation}, we demonstrate the correlation between measured and predicted shear depths of our four fit models.
For a motion-planning cost function, it is more important that the correlation points are ordered (i.e., increasing) rather than lying along the diagonal $y=x$.
For all four models, we qualitatively see a strong correlation between measured and predicted shear depths.

%
%

\begin{figure}[tpb]
  \centering
  {
    \def\svgwidth{\columnwidth}
    \small
    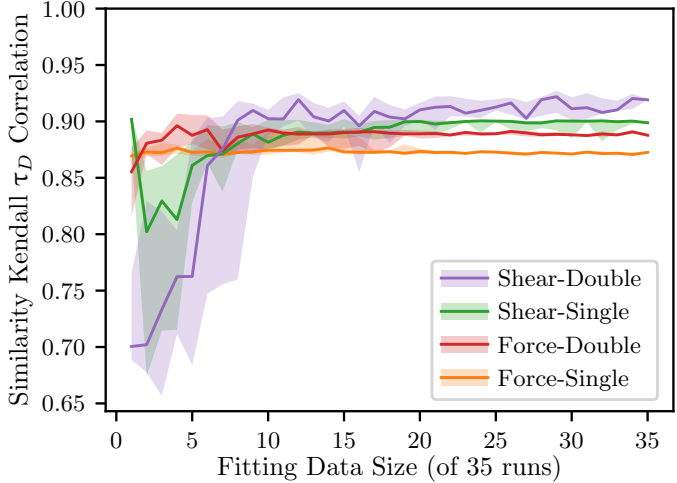
  }
  \vspace{-1em}
  \caption[%
    Fitting data size versus similarity Kendall $\uptau$ correlation coefficient.
  ]{%
    Fitting data size versus similarity Kendall $\uptau$ correlation coefficient.
    For each fitting data size, $d$, we performed ten separate fits using $d$ randomly chosen runs from our full dataset of \num{35} runs, with the remainder kept as a hold-out dataset.
    The similarity Kendall $\uptau$ correlation is computed over all \num{35} runs.
    From the ten fits at each fitting data size, we show the median with a solid line between upper and lower quartiles in the shaded regions.
    A perfect ordering is correlation \num{1} and a random ordering in expectation is correlation \num{0}.
  }
  \label{fig:multifit-tau}
\end{figure}

%
%
To determine the effect of the size of dataset used in model fitting on the ranking capability of our four models, we compare the obtained $\tau_D$ for various fitting dataset sizes in Fig.~\ref{fig:multifit-tau}.
The \ModelForceSingle, \ModelForceDouble, \ModelShearSingle, and \ModelShearDouble models converge to a median $\uptau_D$ of \num{0.87}, \num{0.89}, \num{0.91}, and \num{0.92}, respectively; with $p$-values between \numrange{1.5e-7}{5.8e-7}.
The \ModelForceSingle, \ModelForceDouble, and \ModelShearSingle appear to converge after a dataset size of \num{2}, \num{10}, and \num{18}, respectively.
The \ModelShearDouble model shows higher variability, but roughly converges at a dataset size of \num{20}.
The $F_{\mathrm{ins}}$-fit models converge much sooner, primarily due to the large number of force samples (approximately \numrange{100}{300} samples per run), compared to a single shearing-depth sample per run.
But the shear-depth-fit models perform better in ranking, primarily because the ranking is also performed on shearing depth.
We also see the double-parameter models can fit better than single-parameter models but require more data.
However, we show that all fit models provide similar discriminatory abilities, some of which require significantly fewer experimental runs to fit well.

\subsection{Motion-Planning Evaluation}

%
%
As our method is the first steerable needle motion planner to explicitly consider the bottleneck cost associated with maximum tissue normal force, we do not have a natural comparison method.
As such, we instead provide evidence that a popular cost function in steerable-needle motion planning, path length, does not produce plans with small maximum tissue normal force values on average.
To do so, we implement a version of our motion planner that minimizes path length.
We refer to the planners that minimize path length and our force cost as the length planner and force planner, respectively.
Our force cost function uses the \ModelShearDouble model parameters from Table~\ref{tbl:models}.

Both planners are implemented with Alg.~\ref{alg:planner} but using different underlying cost functions.
We set a timeout of \SI{100}{\s} and an approximation factor $\varepsilon = 0.0001$.
We implement random restarts if no solution is found after \num{10 000} RG-RRT iterations (i.e., number of sampled needle tip positions), increasing this limit by \SI{5}{\percent} each time we perform a random restart.
On each run, both planners are provided the same random seed, ensuring their first solutions are identical before path optimization begins.

\begin{figure*}[tpb]
  \centering
  {
    \def\svgwidth{0.7\textwidth}
    \Large
    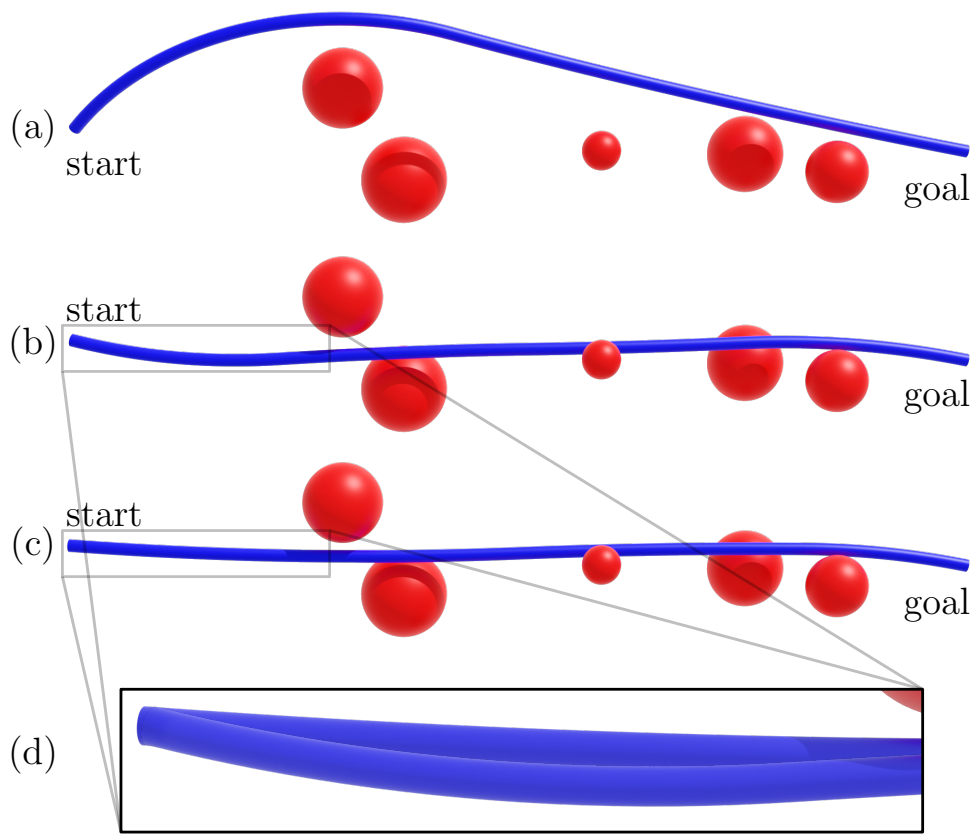
  }
  \caption[%
    Generated collision-free trajectories in a randomly generated 3-D environment with spherical obstacles.
  ]{%
    Generated collision-free trajectories in a randomly generated 3-D environment with spherical obstacles.
    Utilizing our force model during motion planning produces paths that reduce the normal force on the surrounding tissue when compared with utilizing length as the cost during planning.
    \textbf{(a)}~Initial solution by both planners before optimization, with maximum tissue normal force of \SI{47.0}{\mN\per\mm} (note, this fit force model predicts shearing begins at \SI{17.6}{\mN\per\mm}).
    %
    \textbf{(b)}~Solution of the planner using the path-length cost, resulting in a maximum tissue normal force of \SI{20.0}{\mN\per\mm}.
    \textbf{(c)}~Solution of our planner using our maximum tissue normal force bottleneck cost, resulting in a maximum tissue normal force of \SI{3.5}{\mN\per\mm}.
    \textbf{(d)}~3$\times$ zoomed-in view of both solutions at the start; the path from (b) is below and (c) is above, with much more curvature on path (b).
    The length difference between (b) and (c) is less than \SI{0.2}{\percent}, but the force cost from the length planner is \num{5.7} times larger than the force planner's force cost, and is beyond the fit force model's average shearing force.
  }
  \vspace{-1em}
  \label{fig:needle-path-spheres}
\end{figure*}

%
%
As shown in Fig.~\ref{fig:needle-path-spheres}, we task the planner with finding a path in 3-D from the start to the goal, where the planner must avoid spherical obstacles.
We evaluate the length and force planners in \num{400} environments with randomly generated spherical obstacles, using a needle of \SI{1}{\mm} OD and a minimum clearance of \SI{1}{\mm}.
The start position and goal pose are identical and fixed for all environments.
We generate the spherical obstacles with the radius sampled uniformly from  \SIrange{2}{10}{\mm} and the center sampled uniformly from the workspace.
Spheres are rejected if they contain the start or goal positions.
We generate these environments with at least four spheres, then continue adding spheres while the trivial motion-planning solution is collision-free.
The trivial solution is a single constant-curvature path between the start and goal.
%

\subsubsection{Results}

Our planner successfully found a plan in \num{331} of the \num{400} randomly generated environments.
Since our environment generation does not ensure the existence of a solution, we discard from our analysis the \num{69} environments with no successful motion plans.

%
%
One example environment is demonstrated in Fig.~\ref{fig:needle-path-spheres}.
The initial path found by both planners (Fig.~\ref{fig:needle-path-spheres}(a)) is suboptimal in both path-length and force costs.
The subsequent paths found by the length planner (Fig.~\ref{fig:needle-path-spheres}(b)) and the force planner (Fig.~\ref{fig:needle-path-spheres}(c)) show a similarly obtained path length (difference of \SI{0.2}{\percent}), but the length planner's path shows significantly more curving near the start (see Fig.~\ref{fig:needle-path-spheres}(d)), which results in \num{5.7} times higher force cost than the force planner's force cost.

\begin{figure*}[tb]
  \centering
  \def\svgwidht{0.7\textwidth}
  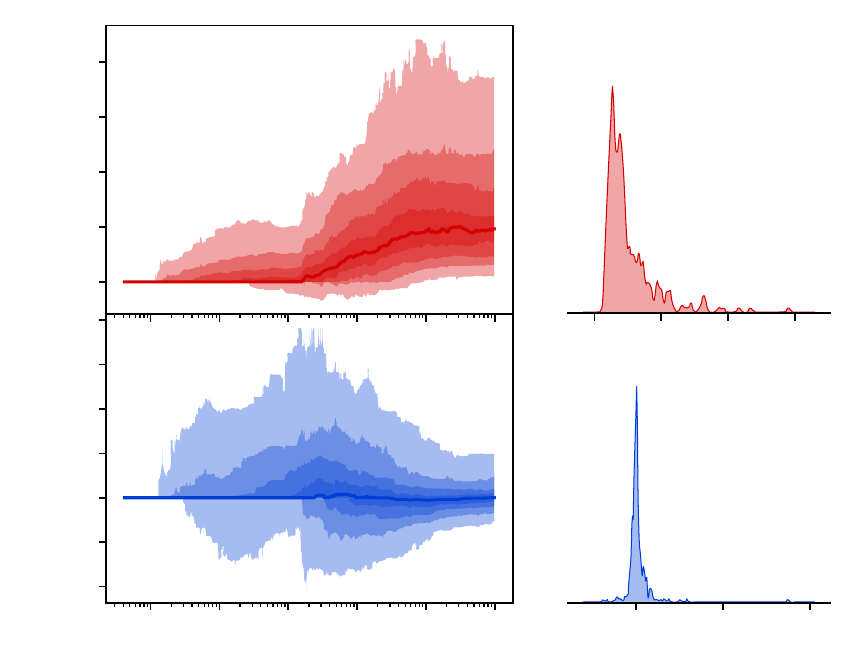
  \caption[%
    From \num{331} random sphere environments, we compare the cost ratios of the length and force planners for both length and force costs versus planning time.
  ]{%
    From \num{331} random sphere environments, we compare the cost ratios of the length and force planners for both length and force costs versus planning time.
    (Top)
    The ratio of force costs with the length planner over the force planner, shown in red.
    A 1.0 ratio (shown as a dotted line) represents the force cost obtained from the force planner.
    (Bottom)
    The ratio of length costs with the force planner over the length planner, shown in blue.
    Likewise, the 1.0 line represents the length planner's achieved length cost.
    The ratios are generated pair-wise over plans for corresponding environments.
    (Left)
    We show the median (solid line) and four quantiles above and below (shaded) \SI{10}{\percent} quantile steps.
    (Right)
    We show the kernel-density estimate (KDE) of the cost ratios at \SI{100}{\s}.
    }
  \label{fig:sphere-path-cost}
\end{figure*}

%
%
As further evidence that the length cost function does not adequately minimize forces, the length planner's returned force costs after \SI{100}{\s} are \num{1.6 +- 1.9} times larger than those from the force planner (i.e., a force ratio of \num{2.6 +- 1.9}).
In Fig.~\ref{fig:sphere-path-cost}, the quantile plot in the upper-left shows that at around \SI{10}{\s}, over \SI{90}{\percent} of the returned paths have a larger force than those returned from the force planner.
The top-right density plot in Fig.~\ref{fig:sphere-path-cost} additionally shows a large spread of force cost ratios after \SI{100}{\s} of planning and shows many plans that exhibit up to ten times the force cost than the force planner.

%
%
We also compare the path lengths over time for each of the planners.
Notably, the force planner produces paths of comparable length to the length planner, but does so indirectly (see~\eqref{eqn:n-solved}) while significantly improving the tissue forces of the paths.
Fig.~\ref{fig:sphere-path-cost} shows the length planner performs slightly better than the force planner in path length.
The length cost ratio between the force-planner and length-planner is on average \num{1.0007} with standard deviation {0.0065} ($p$-value of \num{0.03} against the null hypothesis of \num{1.0} mean).
This difference is imperceptible and well within the standard deviation of path lengths from the length planner (\SI{1.0}{\percent}).
This shows that optimizing for tissue forces enables found paths with drastically less force on surrounding tissue without measurably sacrificing in path length.

It is an intuitive result that a planner optimizing for tissue normal forces would produce plans that have lower tissue normal forces than one that was not.
However we present this analysis to demonstrate that path length is not a sufficient proxy metric for tissue normal forces, even though path length has an impact on tissue normal forces as shown in~\eqref{eqn:n-solved}.
The intuition for this is shown in Fig.~\ref{fig:tissue-force-heatmap}(c).
Trajectories that are identical both in length and maximum curvature can have dramatically different maximum tissue normal forces.
This highlights the need for considering the tissue normal forces explicitly during motion-planning.

\subsection{Anatomical Environment}

We next demonstrate initial feasibility of using the motion-planning method that considers tissue normal forces with a clinically relevant task in an anatomical environment.
We task the motion planner with planning a path for a needle from a patient's chest wall to a target deep in the lung, as in percutaneous lung tumor biopsy.
We utilize a CT scan from the 2017 lung CT segmentation challenge~\cite{Yang2018_MP,Yang2017_TCIA} in the cancer imaging archive~\cite{Clark2013_JDI}.
Using the segmentation method of Fu \etal~\cite{Fu2018_IROS} we segment the large vasculature and bronchial trees in the lung.
These are used as obstacles for the motion planner that must be avoided.

\begin{figure*}[tb]
  \centering
  {
    \def\svgwidth{0.8\textwidth}
    \LARGE
\begingroup%
  \makeatletter%
  \providecommand\color[2][]{%
    \errmessage{(Inkscape) Color is used for the text in Inkscape, but the package 'color.sty' is not loaded}%
    \renewcommand\color[2][]{}%
  }%
  \providecommand\transparent[1]{%
    \errmessage{(Inkscape) Transparency is used (non-zero) for the text in Inkscape, but the package 'transparent.sty' is not loaded}%
    \renewcommand\transparent[1]{}%
  }%
  \providecommand\rotatebox[2]{#2}%
  \newcommand*\fsize{\dimexpr\f@size pt\relax}%
  \newcommand*\lineheight[1]{\fontsize{\fsize}{#1\fsize}\selectfont}%
  \ifx\svgwidth\undefined%
    \setlength{\unitlength}{252.01721912bp}%
    \ifx\svgscale\undefined%
      \relax%
    \else%
      \setlength{\unitlength}{\unitlength * \real{\svgscale}}%
    \fi%
  \else%
    \setlength{\unitlength}{\svgwidth}%
  \fi%
  \global\let\svgwidth\undefined%
  \global\let\svgscale\undefined%
  \makeatother%
  \begin{picture}(1,0.67384859)%
    \lineheight{1}%
    \setlength\tabcolsep{0pt}%
    \put(0,0){\includegraphics[width=\unitlength,page=1]{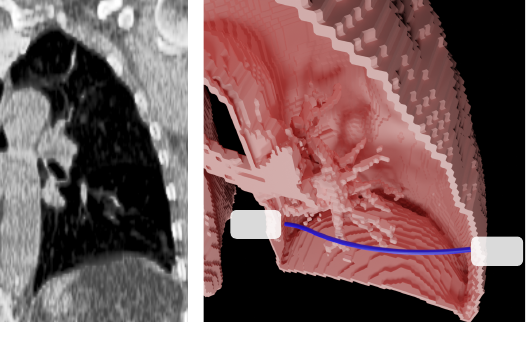}}%
    \put(0.1790033,0.00933486){\color[rgb]{0,0,0}\makebox(0,0)[t]{\lineheight{1.25}\smash{\begin{tabular}[t]{c}(a)\end{tabular}}}}%
    \put(0.69388325,0.00933486){\color[rgb]{0,0,0}\makebox(0,0)[t]{\lineheight{1.25}\smash{\begin{tabular}[t]{c}(b)\end{tabular}}}}%
    \put(0.94572853,0.18269933){\color[rgb]{0,0,0}\makebox(0,0)[t]{\lineheight{1.25}\smash{\begin{tabular}[t]{c}start\end{tabular}}}}%
    \put(0.48701026,0.23597143){\color[rgb]{0,0,0}\makebox(0,0)[t]{\lineheight{1.25}\smash{\begin{tabular}[t]{c}goal\end{tabular}}}}%
  \end{picture}%
\endgroup%

  }
  \caption[%
    We demonstrate the feasibility of utilizing our force-based cost function in motion planning for a steerable needle in an anatomically relevant scenario.
  ]{%
    We demonstrate the feasibility of utilizing our force-based cost function in motion planning for a steerable needle in an anatomically relevant scenario.
    \textbf{(a)} The left lung from one slice from the original CT scan where the segmentation originates.
    \textbf{(b)} We show an example generated plan (blue) by our motion planner for a needle insertion site at the boundary of a patient's lung near the chest wall to reach a target deep in the lung.
    The plan must curve around the obstacle in the lung while minimizing tissue normal forces.
    The shown segmentation includes the chest wall and large bronchial tubes with a cutout to aid in visualization.
    The plan passes close to an anatomical obstacle but curves around it while minimizing the tissue normal forces.
  }
  \label{fig:path-in-anatomy}
\end{figure*}

We ran the force and length planners on this problem \num{100} times with a timeout of \num{100} seconds each.
We show in Fig.~\ref{fig:path-in-anatomy} one example plan from our force planner; demonstrating its ability to find a path from the start to the goal while avoiding the obstacles and doing so while minimizing tissue normal forces.

The force cost ratio between the length and force planners was \numquart{1.01}{0.893}{1.16}---a wide spread---yet the length cost ratio between the force and length planners was only \numquart{1.0001}{0.997}{1.003}.

  \section{Conclusion}
\label{sec:conclusion}

This work provides the following contributions:
(i) a simple and efficient Cosserat-string-based needle-to-tissue force and friction model,
(ii) a clinically motivated and computationally efficient motion-planning cost function for needle steering based on our needle-to-tissue force and friction model, and a motion-planning algorithm that leverages our force-based cost function, and
(iii) two effective strategies for fitting our force model parameters with easily obtained insertion force measurements or labeled shearing events, depending on whichever is easier to obtain.
Minimizing the tissue normal forces during needle steering has the potential to significantly reduce the risk of critical tissue damage events, thus improving patient outcomes.




In this work, we utilized a homogeneous gel phantom to validate and fit our force model.
In future work, we intend to utilize real heterogeneous tissues involved in clinical procedures to evaluate our method's efficacy in real clinical settings.
Shearing events utilized in the model fitting process would then need to be measured in ex vivo tissue using medical imaging, such as fluoroscopy.

%

This work's analysis does not yet consider the stochasticity in needle control~\cite{Park2010_ICRA} and in damage outcome.
We intend to investigate the use of our cost function in stochastic risk metrics for risk-based planning under uncertainty~\cite{Majumdar2020_ISRR}.
First, we wish to reformulate our trajectory cost in probabilistic terms, directly modeling the probability of tissue damage using our model's predicted forces.
Second, we plan to incorporate stochasticity in robot control in an uncertainty-aware planning framework such as Partially Observable Markov Decision Processes (POMDP)~\cite{Kurniawati2022_CRAS}.
This will enable us to take a principled approach to trajectory risk and employ a well-behaved risk metric such as Conditional Value at Risk (CVaR)~\cite{Majumdar2020_ISRR}.



  \section*{Acknowledgment}
The authors would like to thank Ron Alterovitz and his group for their insights and assistance with segmentation, and Robert J. Webster III and his group for insights and valuable discussions.

This research was supported in part by the U.S. National Science Foundation (NSF) under Awards IIS-1652588 (CAREER) and CMMI-2133027, by the Israeli Ministry of Science \& Technology grants no. 3-16079 and 3-17385 and  in part by the United States-Israel Binational Science Foundation (BSF) grants no. 2019703 and 2021643

  \bibliographystyle{ws-jmrr}
  \bibliography{references}

\begin{thebibliography}{10}

\bibitem{Abolhassani2007_MEP}
N.~Abolhassani, R.~Patel and M.~Moallem, Needle insertion into soft tissue: A
  survey, {\em Med. Eng. Phys.} {\bf 29} (May 2007)  413--431.

\bibitem{Reed2011_RAM}
K.~B. Reed, A.~Majewicz, V.~Kallem, R.~Alterovitz, K.~Goldberg, N.~J. Cowan and
  A.~M. Okamura, Robot-assisted needle steering, {\em IEEE Robot. Autom. Mag.}
  {\bf 18} (December 2011)  35--46.

\bibitem{Webster2006_IJRR}
R.~J. Webster, III, J.~S. Kim, N.~J. Cowan, G.~S. Chirikjian and A.~M. Okamura,
  Nonholonomic modeling of needle steering, {\em Int. J. Rob. Res.} {\bf 25}
  (May 2006)  509--525.

\bibitem{Rox2020_IA}
M.~Rox, M.~Emerson, T.~E. Ertop, I.~Fried, M.~Fu, J.~Hoelscher, A.~Kuntz,
  J.~Granna, J.~E. Mitchell, M.~Lester, F.~Maldonado, E.~A. Gillaspie, J.~A.
  Akulian, R.~Alterovitz and R.~J. Webster, III, Decoupling steerability from
  diameter: Helical dovetail laser patterning for steerable needles, {\em IEEE
  Access} {\bf 8} (October 2020)  181411--181419.

\bibitem{vandeBerg2014_TMECH}
N.~J. {van de Berg}, D.~J. {van Gerwen}, J.~Dankelman and J.~J. {van den
  Dobbelsteen}, Design choices in needle steering\textemdash{{A}} review, {\em
  IEEE/ASME Trans. Mechatron.} {\bf 20} (October 2014)  2172--2183.

\bibitem{Yang2018_JMRR}
F.~Yang, M.~Babaiasl and J.~P. Swensen, Fracture-directed steerable needles,
  {\em J. Med. Robot. Res.} {\bf 4} (March 2018).

\bibitem{Patil2014_TRO}
S.~Patil, J.~Burgner, R.~J. Webster, III and R.~Alterovitz, Needle steering in
  3-{{D}} via rapid replanning, {\em IEEE Trans. Robot.} {\bf 30} (August 2014)
   853--864.

\bibitem{Kuntz2015_IROS}
A.~Kuntz, L.~G. Torres, R.~H. Feins, R.~J. Webster, III and R.~Alterovitz,
  Motion planning for a three-stage multilumen transoral lung access system,
  {\em 2015 {{IEEE Int}}. {{Conf}}. {{Intell}}. {{Robots Syst}}. ({{IROS}})\/},
   {Hamburg, Germany} (September 2015), pp. 3255--3261.

\bibitem{Fu2018_IROS}
M.~Fu, A.~Kuntz, R.~J. Webster, III and R.~Alterovitz, Safe motion planning for
  steerable needles using cost maps automatically extracted from pulmonary
  images, {\em 2018 {{IEEE Int}}. {{Conf}}. {{Intell}}. {{Robots Syst}}.
  ({{IROS}})\/},  {Madrid, Spain} (October 2018), pp. 4942--4949.

\bibitem{Pinzi2019_IJCARS}
M.~Pinzi, S.~Galvan and F.~{Rodriguez~y~Baena}, The adaptive {{Hermite}}
  fractal tree ({{AHFT}}): A novel surgical {{3D}} path planning approach with
  curvature and heading constraints, {\em Int. J. Comput. Assist. Radiol.
  Surg.} {\bf 14} (April 2019)  659--670.

\bibitem{Majewicz2012_TBME}
A.~Majewicz, S.~P. Marra, M.~G. van Vledder, M.~Lin, M.~A. Choti, D.~Y. Song
  and A.~M. Okamura, Behavior of tip-steerable needles in ex vivo and in vivo
  tissue, {\em IEEE Trans. Biomed. Eng.} {\bf 59} (October 2012)  2705--2715.

\bibitem{Alterovitz2005_ICRA}
R.~Alterovitz, K.~Goldberg and A.~M. Okamura, Planning for steerable bevel-tip
  needle insertion through {{2D}} soft tissue with obstacles, {\em Proc. 2005
  {{IEEE Int}}. {{Conf}}. {{Robot}}. {{Autom}}.\/},  {Barcelona, Spain} (April
  2005), pp. 1652--1657.

\bibitem{Minhas2009_EMBS}
D.~Minhas, J.~A. Engh and C.~N. Riviere, Testing of neurosurgical needle
  steering via duty-cycled spinning in brain tissue in vitro, {\em 2009
  {{Annu}}. {{Int}}. {{Conf}}. {{IEEE Eng}}. {{Med}}. {{Biol}}. {{Soc}}.\/},
  (September 2009), pp. 258--261.

\bibitem{Swaney2017_JMRR}
P.~J. Swaney, A.~W. Mahoney, B.~I. Hartley, A.~A. Remirez, E.~Lamers, R.~H.
  Feins, R.~Alterovitz and R.~J. Webster, III, Toward transoral peripheral lung
  access: Combining continuum robots and steerable needles, {\em J. Med. Robot.
  Res.} {\bf 2} (March 2017).

\bibitem{Engh2006_EMBS}
J.~A. Engh, G.~Podnar, D.~Kondziolka and C.~N. Riviere, Toward effective needle
  steering in brain tissue, {\em 2006 {{Int}}. {{Conf}}. {{IEEE Eng}}. {{Med}}.
  {{Biol}}. {{Soc}}. {{EMBC}}\/},  {New York, NY, USA} (September 2006), pp.
  559--562.

\bibitem{Swaney2012_TBME}
P.~J. Swaney, J.~Burgner, H.~B. Gilbert and R.~J. Webster, III, A flexure-based
  steerable needle: High curvature with reduced tissue damage, {\em IEEE Trans.
  Biomed. Eng.} {\bf 60} (November 2012)  906--909.

\bibitem{Bui2016_IME}
V.~K. Bui, S.~Park, J.-O. Park and S.~Y. Ko, A novel curvature-controllable
  steerable needle for percutaneous intervention, {\em Proc. Inst. Mech. Eng.
  Part H: J. Eng. Med.} {\bf 230} (August 2016)  727--738.

\bibitem{Gerboni2017_RAL}
G.~Gerboni, J.~D. Greer, P.~F. Laeseke, G.~L. Hwang and A.~M. Okamura, Highly
  articulated robotic needle achieves distributed ablation of liver tissue,
  {\em IEEE Robot. Autom. Lett.} {\bf 2} (July 2017)  1367--1374.

\bibitem{vandeBerg2015_MEP}
N.~J. {van de Berg}, J.~Dankelman and J.~J. {van den Dobbelsteen}, Design of an
  actively controlled steerable needle with tendon actuation and {{FBG-based}}
  shape sensing, {\em Med. Eng. Phys.} {\bf 37} (June 2015)  617--622.

\bibitem{Frasson2012_JRS}
L.~Frasson, F.~Ferroni, S.~Y. Ko, G.~Dogangil and F.~{Rodriguez y Baena},
  Experimental evaluation of a novel steerable probe with a programmable bevel
  tip inspired by nature, {\em J. Robot. Surg.} {\bf 6} (June 2012)  189--197.

\bibitem{Schwehr2022_HSMR}
T.~J. Schwehr, A.~J. Sperry, J.~D. Rolston, M.~D. Alexander, J.~J. Abbott and
  A.~Kuntz, Toward targeted therapy in the brain by leveraging screw-tip soft
  magnetically steerable needles, {\em Proc. 14th {{Hamlyn Symp}}. {{Med}}.
  {{Robot}}. 2022\/},  {London, UK} (June 2022), pp. 81--82.

\bibitem{Ilami2020_NSR}
M.~Ilami, R.~J. Ahmed, A.~Petras, B.~Beigzadeh and H.~Marvi, Magnetic needle
  steering in soft phantom tissue, {\em Sci. Rep.} {\bf 10} (February 2020).

\bibitem{Hong2020_TBME}
A.~Hong, A.~J. Petruska, A.~Zemmar and B.~J. Nelson, Magnetic control of a
  flexible needle in neurosurgery, {\em IEEE Trans. Biomed. Eng.} {\bf 68}
  (February 2021)  616--627.

\bibitem{Gessert2019_IJCARS}
N.~Gessert, T.~Priegnitz, T.~Saathoff, S.-T. Antoni, D.~Meyer, M.~F. Hamann,
  K.-P. J{\"u}nemann, C.~Otte and A.~Schlaefer, Spatio-temporal deep learning
  models for tip force estimation during needle insertion, {\em Int. J. Comput.
  Assist. Radiol. Surg.} {\bf 14} (September 2019)  1485--1493.

\bibitem{Gidde2020_BioinspirBiomim}
S.~T.~R. Gidde, A.~Ciuciu, N.~Devaravar, R.~Doracio, K.~Kianzad and P.~Hutapea,
  Effect of vibration on insertion force and deflection of bioinspired needle
  in tissues, {\em Bioinspir. Biomim.} {\bf 15} (July 2020).

\bibitem{Tsumura2019_JMRR}
R.~Tsumura, Y.~Takishita and H.~Iwata, Needle insertion control method for
  minimizing both deflection and tissue damage, {\em J. Med. Robot. Res.} {\bf
  4} (March 2019).

\bibitem{Webster2005_ICRA}
R.~J. Webster, III, J.~Memisevic and A.~M. Okamura, Design considerations for
  robotic needle steering, {\em Proc. 2005 {{IEEE Int}}. {{Conf}}. {{Robot}}.
  {{Autom}}.\/},  {Barcelona, Spain} (April 2005), pp. 3599--3605.

\bibitem{Oldfield2013_CMBBE}
M.~Oldfield, D.~Dini, G.~Giordano and F.~{Rodriguez y Baena}, Detailed finite
  element modelling of deep needle insertions into a soft tissue phantom using
  a cohesive approach, {\em Comput. Methods Biomech. Biomed. Engin.} {\bf 16}
  (May 2013)  530--543.

\bibitem{Takabi2017_MEP}
B.~Takabi and B.~L. Tai, A review of cutting mechanics and modeling techniques
  for biological materials, {\em Med. Eng. Phys.} {\bf 45} (July 2017)  1--14.

\bibitem{Antman2005_Chapter14}
S.~S. Antman, {\em Problems in Nonlinear Elasticity}, {\em Nonlinear Problems
  of Elasticity\/}, Applied {{Mathematical Sciences}}, Vol.~107 ({Springer},
  {New York, NY, USA}, 2005), {New York, NY, USA}, ch.~14, pp. 513--584, second
  edn.

\bibitem{Salzman2019_CACM}
O.~Salzman, Sampling-based robot motion planning, {\em Commun. ACM} {\bf 62}
  (September 2019)  54--63.

\bibitem{LaValle2001_IJRR}
S.~M. LaValle and J.~J. Kuffner, Randomized kinodynamic planning, {\em Int. J.
  Rob. Res.} {\bf 20} (May 2001)  378--400.

\bibitem{Kavraki1996_TRA}
L.~E. Kavraki, P.~{\v S}vestka, J.-C. Latombe and M.~H. Overmars, Probabilistic
  roadmaps for path planning in high-dimensional configuration spaces, {\em
  IEEE Trans. Robot. Autom.} {\bf 12} (August 1996)  566--580.

\bibitem{Yong2004_CAGD}
J.-H. Yong and F.~F. Cheng, Geometric {{Hermite}} curves with minimum strain
  energy, {\em Comput. Aided Geom. Des.} {\bf 21} (March 2004)  281--301.

\bibitem{Fu2021_RSS}
M.~Fu, O.~Salzman and R.~Alterovitz, Toward certifiable motion planning for
  medical steerable needles, {\em Robot. {{Sci}}. {{Syst}}. {{XVII}}\/},   {\bf
  2021}, {Virtual} (July 2021).

\bibitem{Fu2022_ICRA}
M.~Fu, K.~Solovey, O.~Salzman and R.~Alterovitz, Resolution-optimal motion
  planning for steerable needles, {\em 2022 {{Int}}. {{Conf}}. {{Robot}}.
  {{Autom}}. {{ICRA}}\/},  {Philadelphia, PA, USA} (May 2022), pp. 9652--9659.

\bibitem{Favaro2018_ICRA}
A.~Favaro, L.~Cerri, S.~Galvan, F.~{Rodriguez y Baena} and E.~De~Momi,
  Automatic optimized {{3D}} path planner for steerable catheters with
  heuristic search and uncertainty tolerance, {\em 2018 {{IEEE Int}}. {{Conf}}.
  {{Robot}}. {{Autom}}. {{ICRA}}\/},  {Brisbane, QLD, Australia} (May 2018),
  pp. 9--16.

\bibitem{Gammell2015_ICRA}
J.~D. Gammell, S.~S. Srinivasa and T.~D. Barfoot, Batch informed trees
  ({{BIT}}*): Sampling-based optimal planning via the heuristically guided
  search of implicit random geometric graphs, {\em 2015 {{IEEE Int}}. {{Conf}}.
  {{Robot}}. {{Autom}}. {{ICRA}}\/},  {Seattle, WA, USA} (May 2015), pp.
  3067--3074.

\bibitem{Patil2010_BioRob}
S.~Patil and R.~Alterovitz, Interactive motion planning for steerable needles
  in {{3D}} environments with obstacles, {\em 2010 3rd {{IEEE RAS EMBS Int}}.
  {{Conf}}. {{Biomed}}. {{Robot}}. {{Biomechatron}}.\/},  {Tokyo, Japan}
  (September 2010), pp. 893--899.

\bibitem{Attaway1999_ITRS}
S.~W. Attaway, The mechanics of friction in rope rescue, {\em Int. {{Tech}}.
  {{Rescue Symp}}. {{ITRS}} 99\/},   {\bf 7}, {Fort Collins, Colorado, USA}
  (November 1999).

\bibitem{Rucker2011_TRO}
D.~C. Rucker and R.~J. Webster, III, Statics and dynamics of continuum robots
  with general tendon routing and external loading, {\em IEEE Trans. Robot.}
  {\bf 27} (December 2011)  1033--1044.

\bibitem{Minhas2007_EMBS}
D.~S. Minhas, J.~A. Engh, M.~M. Fenske and C.~N. Riviere, Modeling of needle
  steering via duty-cycled spinning, {\em 2007 29th {{Annu}}. {{Int}}.
  {{Conf}}. {{IEEE Eng}}. {{Med}}. {{Biol}}. {{Soc}}.\/},  {Lyon, France}
  (2007), pp. 2756--2759.

\bibitem{Qi2022_TIMC}
Z.~Qi, Q.~Luo and H.~Zhang, A tube-based robust {{MPC}} for duty-cycled
  rotation needle steering systems with bounded disturbances, {\em Trans. Inst.
  Meas. Control} {\bf 44} (February 2022)  960--970.

\bibitem{Branch1999_JSC}
M.~A. Branch, T.~F. Coleman and Y.~Li, A subspace, interior, and conjugate
  gradient method for large-scale bound-constrained minimization problems, {\em
  SIAM J. Sci. Comput.} {\bf 21} (January 1999)  1--23.

\bibitem{Solovey2017_IROS}
K.~Solovey and D.~Halperin, Efficient sampling-based bottleneck pathfinding
  over cost maps, {\em 2017 {{IEEE Int}}. {{Conf}}. {{Intell}}. {{Robots
  Syst}}. ({{IROS}})\/},  {Vancouver, BC, Canada} (September 2017), pp.
  2003--2009.

\bibitem{Holladay2019_RAL}
R.~Holladay, O.~Salzman and S.~Srinivasa, Minimizing task-space {{Fr\'echet}}
  error via efficient incremental graph search, {\em IEEE Robot. Autom. Lett.}
  {\bf 4} (April 2019)  1999--2006.

\bibitem{Niyaz2019_IROS}
S.~Niyaz, A.~Kuntz, O.~Salzman, R.~Alterovitz and S.~S. Srinivasa, Optimizing
  motion-planning problem setup via bounded evaluation with application to
  following surgical trajectories, {\em 2019 {{IEEE Int}}. {{Conf}}.
  {{Intell}}. {{Robots Syst}}. ({{IROS}})\/},   {\bf 2019}, {Macau, China}
  (November 2019), pp. 1355--1362.

\bibitem{Karaman2011_IJRR}
S.~Karaman and E.~Frazzoli, Sampling-based algorithms for optimal motion
  planning, {\em Int. J. Rob. Res.} {\bf 30} (June 2011)  846--894.

\bibitem{Salzman2016_TRO}
O.~Salzman and D.~Halperin, Asymptotically near-optimal {{RRT}} for fast,
  high-quality motion planning, {\em IEEE Trans. Robot.} {\bf 32} (June 2016)
  473--483.

\bibitem{Hauser2016_TRO}
K.~Hauser and Y.~Zhou, Asymptotically optimal planning by feasible kinodynamic
  planning in a state-cost space, {\em IEEE Trans. Robot.} {\bf 32} (December
  2016)  1431--1443.

\bibitem{Kleinbort2020_ICRA}
M.~Kleinbort, E.~Granados, K.~Solovey, R.~Bonalli, K.~E. Bekris and
  D.~Halperin, Refined analysis of asymptotically-optimal kinodynamic planning
  in the state-cost space, {\em 2020 {{IEEE Int}}. {{Conf}}. {{Robot}}.
  {{Autom}}. {{ICRA}}\/},  {Paris, France} (May 2020), pp. 6344--6350.

\bibitem{Kumar2010_WWW}
R.~Kumar and S.~Vassilvitskii, Generalized distances between rankings, {\em
  {{WWW}} '10 {{Proc}}. 19th {{Int}}. {{Conf}}. {{World Wide Web}}\/},
  {Raleigh, NC, USA} (April 2010), pp. 571--580.

\bibitem{Yang2018_MP}
J.~Yang, H.~Veeraraghavan, S.~G. Armato~III, K.~Farahani, J.~S. Kirby,
  J.~{Kalpathy-Kramer}, W.~{van Elmpt}, A.~Dekker, X.~Han and X.~Feng,
  Autosegmentation for thoracic radiation treatment planning: A grand challenge
  at {{AAPM}} 2017, {\em Med. Phys.} {\bf 45} (October 2018)  4568--4581.

\bibitem{Yang2017_TCIA}
J.~Yang, G.~Sharp, H.~Veeraraghavan, W.~{van Elmpt}, A.~Dekker, T.~Lustberg and
  M.~Gooding, Data from lung {{CT}} segmentation challenge (version 3)
  [dataset] (May 2017), \url{https://doi.org/10.7937/K9/TCIA.2017.3R3FVZ08}.

\bibitem{Clark2013_JDI}
K.~Clark, B.~Vendt, K.~Smith, J.~Freymann, J.~Kirby, P.~Koppel, S.~Moore,
  S.~Phillips, D.~Maffitt and M.~Pringle, The cancer imaging archive
  ({{TCIA}}): Maintaining and operating a public information repository, {\em
  J. Digit. Imaging} {\bf 26} (July 2013)  1045--1057.

\bibitem{Park2010_ICRA}
W.~Park, K.~B. Reed, A.~M. Okamura and G.~S. Chirikjian, Estimation of model
  parameters for steerable needles, {\em 2010 {{IEEE Int}}. {{Conf}}.
  {{Robot}}. {{Autom}}.\/},  {Anchorage, AK, USA} (January 2010), pp.
  3703--3708.

\bibitem{Majumdar2020_ISRR}
A.~Majumdar and M.~Pavone, How should a robot assess risk? {{Towards}} an
  axiomatic theory of risk in robotics, {\em Robot. {{Res}}. 18th {{Int}}.
  {{Symp}}. {{ISRR}}\/},  eds. N.~M. Amato, G.~Hager, S.~Thomas and
  M.~{Torres-Torriti} {\em Springer {{Proceedings}} in {{Advanced Robotics}}}
  {\bf 10}, ({Springer Cham}, {Puerto Varas, Chile}, d2020), pp. 75--84.

\bibitem{Kurniawati2022_CRAS}
H.~Kurniawati, Partially observable markov decision processes and robotics,
  {\em Annu. Rev. Control Robot. Auton. Syst.} {\bf 5} (January 2022)
  253--277.

\end{thebibliography}

\end{document}